\documentclass[3p,12pt]{elsarticle}

\usepackage{graphicx}
\usepackage[table,xcdraw]{xcolor}
\usepackage{amsfonts,amsmath}
\usepackage{bbm}
\usepackage{lineno,hyperref}
\usepackage{subfig}
\usepackage{float}
\usepackage{multirow}
\usepackage[section]{placeins} 
\usepackage[bottom]{footmisc}
\usepackage{rotating}

\usepackage{appendix}

\newcommand{\one}{\mathbbm{1}}

\makeatletter
\def\els@aparagraph[#1]#2{\elsparagraph[#1]{#2\@addpunct{.}}}
\def\els@bparagraph#1{\elsparagraph*{#1\@addpunct{.}}}
\makeatother

\begin{document}

\begin{frontmatter}

\title{Pavementscapes: a large-scale hierarchical image dataset for asphalt pavement damage segmentation\footnote{The Pavementscapes dataset is available at \url{https://github.com/tongzheng1992/Pavementscapes}.}}


\author[seu]{Zheng Tong}
\ead{tongzheng@seu.edu.cn}
\author[seu]{Tao Ma\corref{cor1}}
\cortext[cor1]{Corresponding author}
\ead{matao@seu.edu.cn}
\author[seu]{Ju Huyan}
\ead{jhuyan@seu.edu.cn}
\author[seu]{Weiguang Zhang}
\ead{wgzhang@seu.edu.cn}


\address[seu]{School of Transportation, Southeast University, Nanjing 211189, China.}

\begin{abstract}
Pavement damage segmentation has benefited enormously from deep learning. 
However, few current public datasets limit the potential exploration of deep learning in the application of pavement damage segmentation. To address this problem, this study has proposed Pavementscapes, a large-scale dataset to develop and evaluate methods for pavement damage segmentation. Pavementscapes is comprised of 4,000 images with a resolution of $1024 \times 2048$, which have been recorded in the real-world pavement inspection projects with 15 different pavements. A total of 8,680 damage instances are manually labeled with six damage classes at the pixel level. The statistical study gives a thorough investigation and analysis of the proposed dataset. The numeral experiments propose the top-performing deep neural networks capable of segmenting pavement damages, which provides the baselines of the open challenge for pavement inspection. The experiment results also indicate the existing problems for damage segmentation using deep learning, and this study provides potential solutions.
\end{abstract}


\end{frontmatter}


\section{Introduction}
\label{sec:introduction}

Performance of pavement surface decays over time due to various factors, such as traffic volume and weather, and, therefore, the understanding of the deterioration degree is essential to efficient maintenance, which aims to keep and improve the high performance of the surface. Pavement damage, a key characteristic of road surface deterioration degree, is evaluated with three approaches: manual, semi-automatic, and fully automatic.

In the manual approach, investigators walk or slowly drive along the pavement to inspect damages, which is subjective and time-consuming. In the semi-automated one, a fast-moving vehicle automatically collects pavement surface images and an off-line and manual process of damage inspection is then performed in workstations, which is still very time-consuming. Compared to the semi-automated one, the fully automatic approach adopts some technologies of computer vision to perform the inspection using road surface images, which brings the potential of real-time data processing with a low labor cost.

In the fully automatic approach, the application of computer vision on pavement visual inspection involves three main tasks. The first one is \emph{damage recognition} \cite{cha2017deep,lin2010potholes,ye2021convolutional}, also known as \emph{damage classification}, in which an algorithm should indicate the category of each damage present in a 2D or 3D pavement image. Another task is \emph{damage detection} \cite{cha2018autonomous,song2021faster}: bounding boxes identify and locate one or more effective damages in a pavement image. The last one is \emph{damage semantic segmentation} \cite{tong2020pavement,tsai2010critical} which  splits an image into multiple sets of pixels and each set has its individual damage category. These splits, called the \emph{segmentation mask}, are regarded as a simplified but informative representation of the original image. For example, the boundary areas of the segments in a mask provide the position information of pavement damages in an image. Segmentation results can also be used to measure the damage morphology, such as the width and length of a crack. Besides, compared to the results of damage recognition and detection, the ones from damage segmentation are more informative and useful.

The explosive development of deep neural networks \cite{lecun2015deep} brings a large number of outstanding and state-of-the-art methods for semantic segmentation. Many top-performing methods are nowadays extended into the application of pavement damage segmentation and have achieved remarkable success \cite{cha2017deep,cha2018autonomous,tong2018recognition,tong2020pavement,zhang2018deep,zhang2017automated}. The predominant reason for the success is the availability of the large-scale and public datasets, such as ImageNet \cite{deng2009imagenet}, PASCAL VOC \cite{everingham2015pascal}, Cityscapes \cite{cordts2015cityscapes}, and Microsoft COCO \cite{lin2014microsoft}. These datasets exploit the powerful capacity of deep neural networks. Unfortunately, there are a small number of public datasets for pavement damage segmentation \cite{stricker2021road}. This condition has limited the potential exploration of deep learning in the application of pavement damage segmentation. Besides, the lack of publicly available datasets makes the existing algorithms incomparable since they are reported as the state-of-the-art ones only in their own datasets.

This study introduces a large-scale dataset, called \emph{Pavementscapes}, to solve the above-mentioned issue of pavement damage segmentation. The contributions of this study can be summarized as follows:
\begin{enumerate}
	\item Propose a large-scale hierarchical image dataset for asphalt pavement damage segmentation. This dataset can be used to train and test the approaches for pavement damage segmentation, especially for deep neural networks. The proposed dataset consists of 4,000 real-world pavement images annotated in image, block, and pixel levels. For each level, there are six categories of visual pavement damages. The \emph{image-level annotations} defines the damage categories shown in each image. The \emph{block-level annotations} use bounding boxes to identify and locate each damage. In the \emph{pixel-level annotations}, each pixel of a pavement image is labeled as one of the damage categories or background. 
	Besides, all images have a view of top-down shooting, which allows users to measure the accurate morphology of pavement damages using segmentation masks.
	\item Compare the top-performing segmentation methods of computer vision on the proposed dataset. This study investigates the state-of-the-art deep-learning algorithms based on the proposed dataset, which provides the baselines of an open challenge for pavement damage segmentation. The comparison study also indicates the existing issues and potential solutions for the segmentation task, including small damage segmentation, unbalanced training set, and over-confidence in modern neural networks.
	\item Record pavement damages with non-iconic views, which are also known as non-canonical perspectives \cite{palmer1981canonical}). An instance has an iconic view if it is near the center of a digital image. Despite the existing gap in human performance, current algorithms segment damage fairly well on iconic views but struggle to do it otherwise – in the partially occluded and amid clutter \cite{hoiem2012diagnosing}, reflecting the complexity of real-world inspection projects. An ideal model should also perform well in the non-iconic views. Thus, the proposed dataset includes numerous pavement damages in non-iconic views.
\end{enumerate}

The rest of the paper is organized as follows. Section \ref{sec:related} begins with the literature review of the public pavement datasets and state-of-the-art segmentation models based on deep learning. Section \ref{sec:dataset} describes the details of the proposed datasets, followed by the damage segmentation experiments in Section \ref{sec:experiments}. Finally, Section \ref{sec:conclusions} concludes this study.
 
\section{Related work}
\label{sec:related}

\subsection{Pavement damage datasets}
\label{sec:dataset_review}

Recent studies have adopted various machine learning algorithms, especially deep learning for automatic pavement damage segmentation, such as \cite{silva2018concrete,zhang2017automated,zhang2016road}. To develop these algorithms, several datasets of pavement images have been released and the majority of them are summarized in Table \ref{tab:existing_datasets}. Three problems can be found in these public datasets. The first one is the views of the pavement images. Most of these datasets captured the images using smartphones with wide views, such as the most successful one RDD-2020 \cite{arya2021deep}. However, many real-world projects of pavement inspection require a top-down view because pavement maintenance needs information on damage morphology, such as the dimensions of cracks and portholes. A wide view cannot accurately provide morphological information since the damage morphology is distorted in the view. In addition, many datasets present damages with an iconic view, appearing objects in a profile unobstructed near the center of a neatly composed image. However, pavement damages with non-iconic views, such as the incomplete and occluded cracks, are common in real-world inspection projects.

Another problem is the annotations. Table \ref{tab:existing_datasets} indicates that the existing datasets only annotate one or two categories of pavement damages, e.g., crack and pothole. However, asphalt pavement inspection requires several visual damage categories, such as eight in \cite{JTGH202007}. Besides, many datasets only provide image- and block-level annotations, which cannot be used to train and test segmentation models. A few datasets include a small number of images with pixel-level annotations, which still cannot meet the requirement of developing a deep neural network for damage segmentation.
 
The last problem is the baseline algorithms on these datasets. Many datasets still use the results of machine learning proposed thirty years ago as their baselines, even though a few novel deep neural networks have been adopted in some datasets. This phenomenon does not explore the potential of deep learning in the application of pavement damage segmentation.

The three problems motivate us to release a new large-scale hierarchical image dataset for asphalt pavement damage segmentation and analyze the proposed dataset with state-of-the-art deep neural networks.

\begin{table}[]
	\caption{Summary of the existing pavement damage datasets. The algorithms in bold are the deep neural networks. ``Pavementscape'' in the last row is the dataset proposed in this study.}\label{tab:existing_datasets}
	\resizebox{\textwidth}{!}{
	\begin{tabular}{lcclcccccccccl}
		\hline
		\multicolumn{1}{c}{\multirow{2}{*}{Dataset}} & \multirow{2}{*}{Images} & \multirow{2}{*}{Resolution} & \multicolumn{1}{c}{\multirow{2}{*}{Data collection device}} & \multicolumn{2}{c}{View}                              & \multicolumn{4}{c}{Damage category}                                                                           & \multicolumn{3}{c}{Annotation level}                                              & \multicolumn{1}{c}{Baseline algorithms}               \\ \cline{5-13}
		\multicolumn{1}{c}{}                         &                         &                             & \multicolumn{1}{c}{}                                        & Top-down                  & Wide-view                 & Crack                     & Pothole                   & Rut                       & Repair                    & Image                     & Block                    & Pixel                     & \multicolumn{1}{c}{}                                  \\ \hline
		Ouma and Hahn \cite{ouma2017pothole}                   & 75                      & $1080\times 1920$                   & Galaxy S5 cammer                                            & \checkmark &                           &                           & \checkmark &                           &                           &                           &                           & \checkmark & Fuzzy c-means                \\
		CrackIT  \cite{oliveira2014crackit}                     & 84                      & $1536\times 2048$                   & Optical device                                              & \checkmark &                           & \checkmark &                           &                           &                           & \checkmark & \checkmark &                           & K-nearest neighbor                \\
		CFD  \cite{shi2016automatic}                  & 118                     & $480\times 320$                     & Iphone 5                                                    & \checkmark &                           & \checkmark &                           &                           &                           &                           &                           & \checkmark & Random decision forests \\
		CrackTree200 \cite{zou2012cracktree}                           & 206                     & $800\times 600$                     & Area-arry camera                                            & \checkmark &                           & \checkmark &                           &                           &                           &                           &                           & \checkmark & Minimum spanning tree     \\
		SDNET2018  \cite{dorafshan2018sdnet2018}                 & 230                     & $4068\times 3456$                   & 16MP Nikon Digital Camera                                   & \checkmark &                           & \checkmark &                           &                           &                           & \checkmark &                           &                           & \textbf{AlexNet}                                               \\
		Crack500 \cite{yang2019feature}        & 500                     & $2000\times 1500$                   & GoPro 7                                                     & \checkmark &                           & \checkmark &                           &                           &                           &                           &                           & \checkmark & \textbf{FPHBN}                                             \\
		\begin{tabular}[c]{@{}l@{}}CrackDataset \cite{huyan2020cracku}\\ (segmentation part)\end{tabular} & 1205                    & $1280\times 960$                   & Action camera                                                     & \checkmark &                           & \checkmark &                           &                           &                           &                           &            \checkmark    & \checkmark & \textbf{U-net}                                             \\
		
		GAPs v1 \cite{eisenbach2017get}              & 1,969                   & $1920\times 1080$                   & Professional camera                                         &                           & \checkmark & \checkmark & \checkmark &                           & \checkmark & \checkmark &                           &                           & \textbf{ASINVOS net}                                           \\
		GAPs v2 \cite{stricker2019improving}              & 2,468                   & $1920\times 1080$                   & Professional camera                                         &                           & \checkmark & \checkmark & \checkmark &                           & \checkmark &                           & \checkmark &                           & \textbf{ASINVOS net} and \textbf{ResNet34}                              \\
		RDD-2020 \cite{arya2021deep}          & 26,620                  & $720\times 960$                     & LG Nexus 5X cameras                           &                           & \checkmark & \checkmark & \checkmark &                           &                           &                           & \checkmark &                           & \textbf{MobileNet}                                             \\
		Pavementscapes                   & 4,000                   & $2048\times 1024$                   & Professional camera                                         & \checkmark &  & \checkmark & \checkmark & \checkmark & \checkmark & \checkmark & \checkmark & \checkmark & \textbf{Tables \ref{tab:overall_testing}}                                                 \\ \hline
	\end{tabular}}
\end{table}

\subsection{Deep neural networks for pavement damage segmentation}
\label{sec:segmentation_review}

After the success of Long et al. \cite{long2015fully} on semantic segmentation, a large number of deep neural networks have been proposed and achieved the state-of-the-art results. Generally, there are two main directions: convolution- and attention-based deep neural networks.

\begin{figure}
	\centering
	\includegraphics[width=\textwidth]{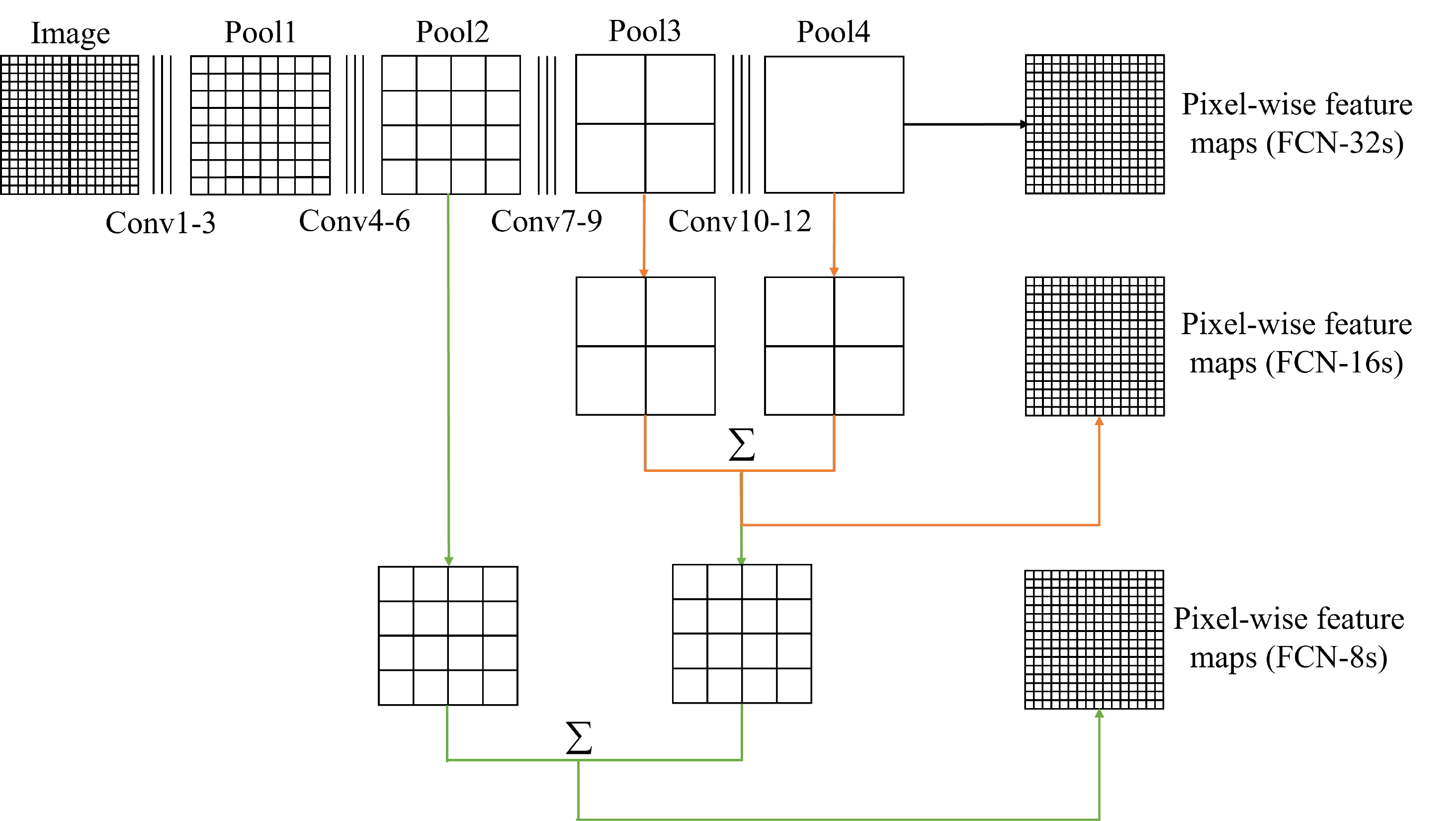}
	\caption{Overview of fully convolution networks \cite{long2015fully}.}\label{fig:FCN}
\end{figure}

\emph{Convolution-based deep neural network}, also known as convolutional neural network (CNN), is a neural network that uses convolution in place of general matrix multiplication in at least one of their layers. For the segmentation problems, the most widely-used CNN architecture is the \emph{fully convolution networks (FCN)} \cite{long2015fully}, as shown in Figure \ref{fig:FCN}, which only consists of locally connected layers, e.g., convolution, pooling, and upsampling layers. No fully connected layer is utilized in this networks. An FCN extracts high-dimension features from a pavement image using convolution and pooling layers and the features are then upsampled into pixel-wise feature maps by upsampling layers. The upsampled feature maps are finally imported into a softmax layer to predict the classes of all pixels in the input image. Many FCN-based models have been used for pavement damage segmentation, such as FCN-8s \cite{tong2020pavement}, U-net \cite{guan2021automated,liu2019computer}, W-Net \cite{han2021crackw}, a series of DeepLab \cite{chen2018encoder,autodeeplab2019}. These studies have demonstrated that the CNN models have significant superiority in pavement damage segmentation, once given enough learning samples with reasonable pixel-wise annotations.

\begin{figure}
	\centering
	\includegraphics[width=\textwidth]{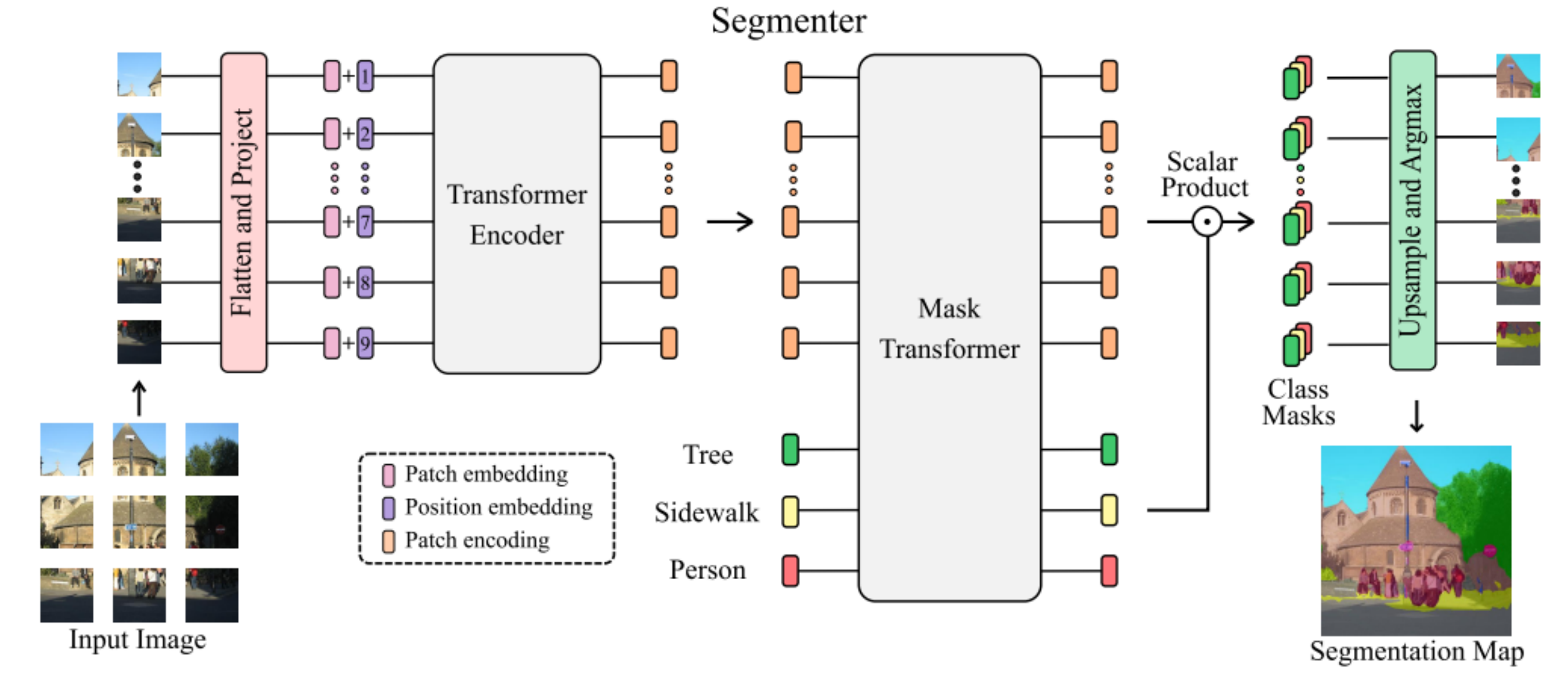}
	\caption{Overview of an attention-based deep neural network \cite{strudel2021segmenter}.}\label{fig:transformer}
\end{figure}

Another direction is the \emph{attention-based deep neural networks}. Compared to an FCN directly using a full image for segmentation, an attention-based network first splits an input into a square-patch grid. Each patch is vectorized by concatenating its channels of all elements and then linearly projected to the required size. After dividing the sample, the network is agnostic to the position information about these patch vectors. Thus, learnable position embeddings are linearly added to each vector, which allows it to learn about the relative or absolute positions of the patches. These embedded patch vectors are then sequentially imported into a transformer encoder. The encoder consists of alternating layers of self-attention and multi-layer perceptron. Self-attention of an embedded patch vector is defined as its relationship to every other vector. Feeding the embedded patch vectors sequentially, a self-attention layer computes their self-attentions as introduced in \cite{dosovitskiy2020image}. These self-attentions are then fed into a multi-layer perceptron layer to handle their dimension. The self-attention outputs of the final transformer encoder are concatenated and imported into a mask transformer. The outputs of the mask transformer, as the pixel-wise feature maps of the input image, are imported into a softmax layer for object segmentation. The processes can be summarized as Figure \ref{fig:transformer}. Even though attention-based networks (e.g., transformer segmentor \cite{strudel2021segmenter}, attention U-net \cite{oktay2018attention} and R2U-net \cite{qin2020u2}) have achieved remarkable results on the majority of benchmark datasets for semantic segmentation, there are only a few applications on pavement damage segmentation \cite{sun2022dma,kang2021efficient,sun2022dma}. Therefore, this study should compare the performances of convolution- and attention-based deep neural networks on pavement damage segmentation, once given a new large-scale hierarchical pavement damage dataset.

\section{Proposed dataset}
\label{sec:dataset}

This section describes the proposed dataset, including image collection (Section \ref{sec:collection}), categories and annotations (Section \ref{sec:class}), the protocol of dataset split (Section \ref{sec:protocol}), and the statistical analysis (Section \ref{sec:statistical}
). The proposed dataset, named \emph{Pavementscapes}, is comprised of 4k images with a resolution of $1024 \times 2048$, which were collected from 15 different pavements in China. A total of 8,680 damage instances in six categories are manually labeled in image, block, and pixel levels.

	\subsection{Image collection}
	\label{sec:collection}
\begin{figure}
    	\centering
    	\includegraphics[width=\textwidth]{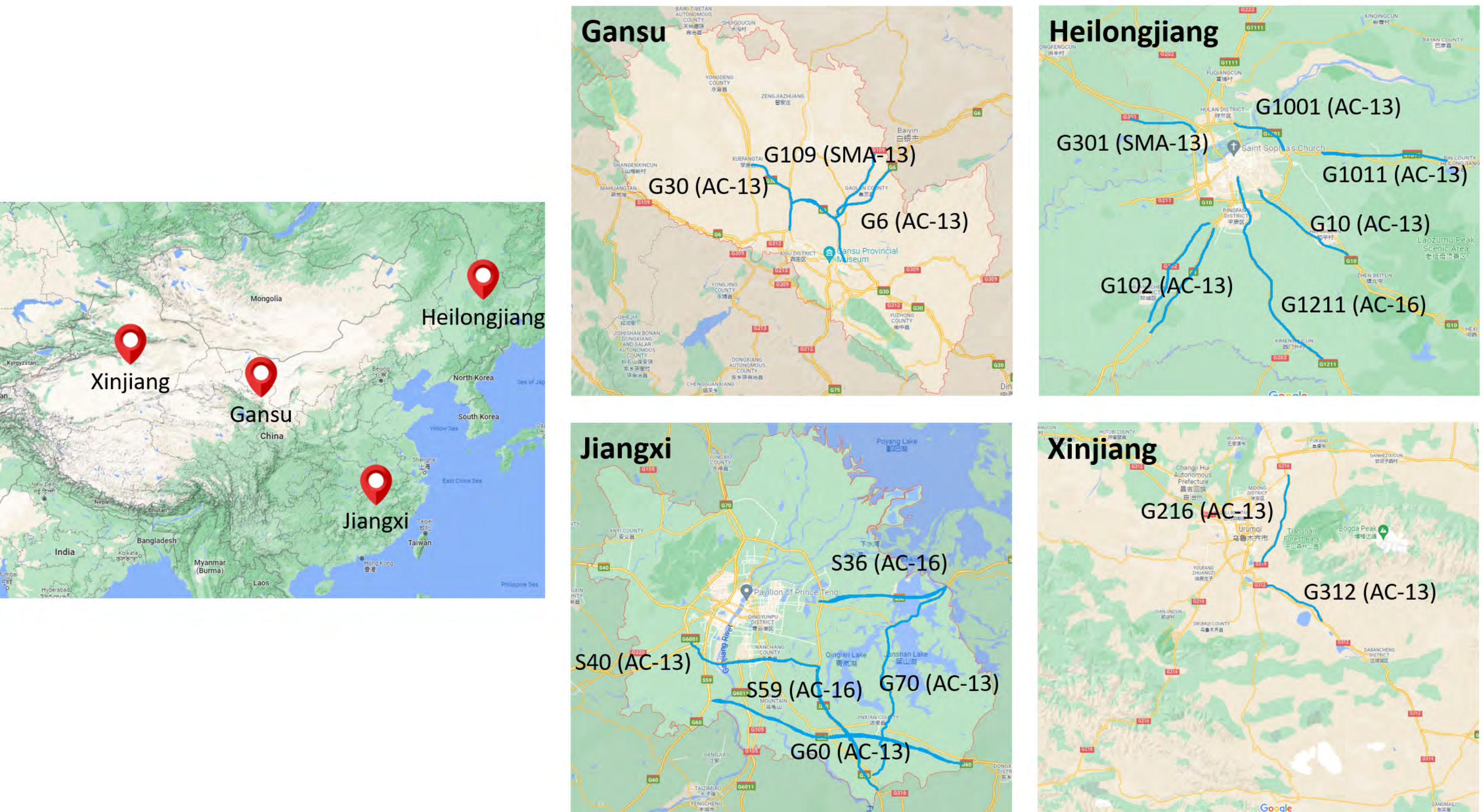}
    	\caption{Study area of the Pavementscape dataset in the Google map. The blue curves are the investigated pavement areas.}\label{fig:study_area}
    \end{figure}

    In order to guarantee the comprehensiveness of the proposed dataset, the areas for image collection are made up of 15 pavements at different locations in China (Jiangxi, Gansu, Heilongjiang, and Xinjiang provinces), which have various service years (1-10 years), traffic volumes, weather, and surface materials (AC-13, AC-16, SMA-13, etc). Figure \ref{fig:study_area} presents the details of these pavements. 
    
\begin{figure}
		\centering
		\includegraphics[width=0.6\textwidth]{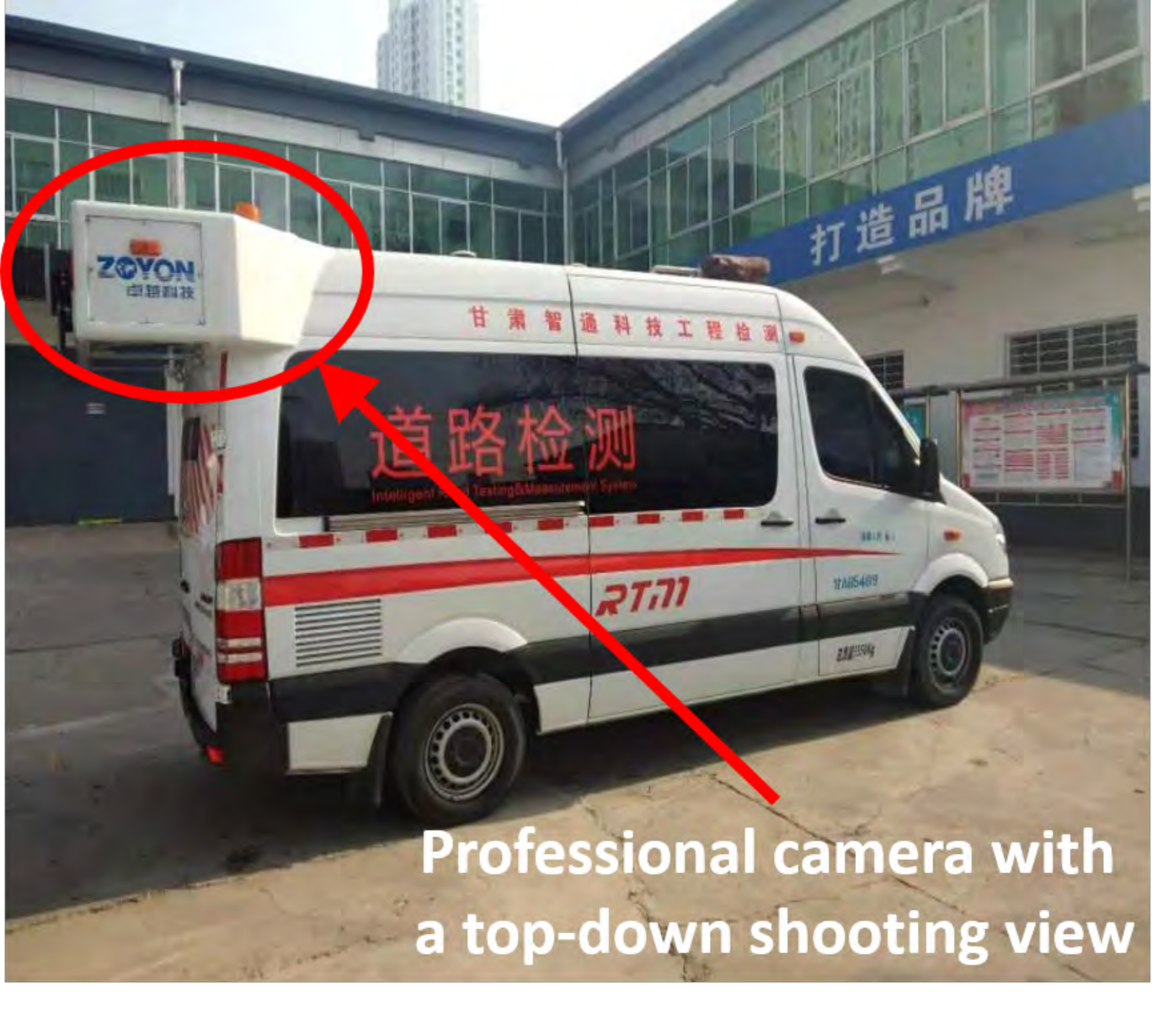}
		\caption{Multi-function detection vehicle equipped with a professional camera.}\label{fig:vehicle}
	\end{figure}

	Pavement images were captured using a multi-function detection vehicle equipped with a professional camera, as shown in Figure \ref{fig:vehicle}. The camera was installed with a top-down shooting view of pavement surfaces, and it captured PNG images with a size of $1024 \times 2048 \times 1$ when the vehicle moved about 60 km/h. More than 500k images were gathered and 4k of them with at least one damage were used to make up the Pavementscapes dataset.
	
    \subsection{Categories and annotations}
    \label{sec:class}
    
    The Pavementscapes dataset consists of six damage categories in total, covering the majority of the visual damage categories in the Chinese Highway Performance Assessment Standards (JTG H20-2007) \cite{JTGH202007}, as shown in Table \ref{tab:dataset_class}.
    
    \begin{table}[]
    	\centering
    	\caption{Pavement damage types in the Pavementscapes dataset and \cite{JTGH202007}.}\label{tab:dataset_class}
    	\begin{tabular}{llc}
    		\hline
    		\multicolumn{2}{l}{Damge type}          & Included in the Pavementscapes dataset or not \\ \hline
    		\multirow{3}{*}{Crack} & Longitudinal   &  \checkmark \\
    		& Lateral        & \checkmark \\
    		& Alligator      & \checkmark \\ \hline
    		\multirow{6}{*}{Others} & Pothole        &     \checkmark \\
    		& Material loose & \texttimes              \\
    		& Rut            & \checkmark \\
    		& Wave crowding  & \texttimes       \\
    		& Repair area    & \checkmark \\ \hline
    	\end{tabular}
    \end{table}

\begin{figure}
		\centering
		\includegraphics[width=\textwidth]{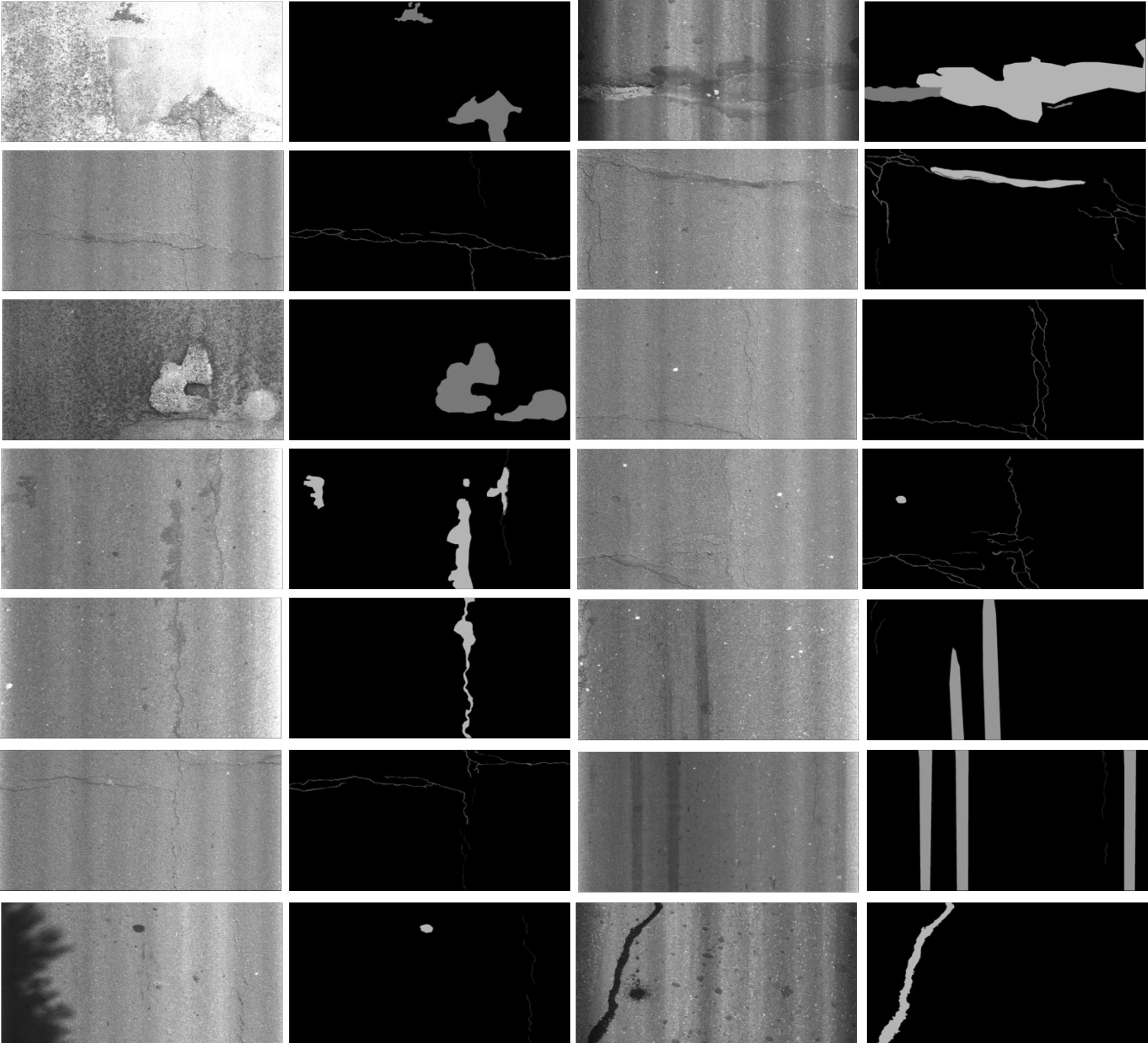}
		\caption{Examples of pavement images and pixel-level annotations from the Pavementscapes dataset. The masks with gray-scale values of 0 are the ``background'' pixels; other masks with different gray-scale values are the pixels belonging to different classes, such that 30, 60, 90, 120, 150, and 180 gray-scale values stand for the pixels of ``longitudinal crack'', ``lateral crack'', ``alligator crack'', ``pothole'', ``rut'', and ``repair area'', respectively.}\label{fig:examples}
	\end{figure}

	The proposed dataset provides annotations at image, block, and pixel levels. Using Labelme library \cite{labelme2016}, these annotations were labeled in-house by five annotators with at least five-year experiments on pavement inspection to guarantee high quality. At the image level, annotators label the damage categories presented in each image, in which multi-class annotations have existed. At the block level, bounding boxes identify and localize damages, which are stored in Excel format. The position information of a bounding box is represented by its left-top and right-bottom coordinates in an image. At the pixel level, annotators assign each pixel to one of the six categories or ``background'' class by labeling these images
	. The ``background'' class has the semantic of ``anything else'' except the six damage categories. The pixel-level annotations are also restored in the PNG format. To guarantee the annotation quality, an annotator should take at least 10 min to label one pavement image. Figure \ref{fig:examples} shows some pavement images and their pixel-level annotations. In practice, these images and their annotations should be transformed into a TFRecord format, which can save computer memory \cite{tensorflow2015-whitepaper}. Even though three types of annotations have been provided, this study only focuses on pixel-level ones. This is because the pixel-level annotations can also represent the information of the image- and block-level ones. The boundary areas of the pixel-level annotations provide the position information of pavement damages in an image, while the pixel classes include the category information of pavement damages. Of course, users can only use the image- and block-level annotations if they only need to train and test a damage detection model.
	
    \subsection{Protocol of dataset split}
    \label{sec:protocol}
    
    The Pavementscapes dataset has been pre-split into separate training, validation, and test sets for any supervised algorithms of computer vision. The protocol of dataset split is not random, but rather in the principle that makes each split representative for various pavement surface scenarios. Specifically, each split set is made up of pavement images collected with the following properties: (i) with a real-world distribution of the damage numbers across individual categories; (ii) in the different geographic locations of China with completely different climate conditions; (iii) at the fine and poor sunshine; (iv) with the different service years. Following this scheme, this study designs a protocol of the Pavemenetscape dataset split with 2,500 training images, 500 validation images, and 1,000 testing images.
    
    To evaluate how representative the three splits w.r.t the protocol properties, an FCN \cite{long2015fully} was trained by the 600 images from the training set and then evaluated by the testing set, as well as eight subsets of the testing set. For each subset, this study randomly selects three-eighths of the testing set. The accuracies of the testing set and its subsets are very uniform, varying less than 2.0\%. Similar phenomena can also be found in the properties of geographic locations and service years. Interestingly, the performance on the fine sunshine is higher than the one on the whole test set. This is mainly because the images in lighting conditions represent damage features better than in the other conditions. To analyze this behavior in-depth, an additional test is performed by using images collected in low- or high-sunshine conditions, observing a 3.2\% accuracy decrease in the low one and a 1.1\% increase in the high one. Similarly, extreme training samples for one condition, such as all training images collected from the SMA pavement, improve the performance on the special testing samples but decrease the one on the whole testing set. These results highlight the comprehensiveness of the proposed dataset that should cover the majority of pavement surface scenes in the real world.
    
    
	\subsection{Statistical analysis}
	\label{sec:statistical}
	
	This section provides a statistical analysis of the proposed datasets, including (i) the distribution of visual damages, (ii) scene complexity under various real-world conditions, (iii) annotation accuracy, and (iv) non-iconic views. Regarding the first aspect, we compare the Pavementscapes dataset to Crack500 \cite{yang2019feature} and CrackDataset \cite{huyan2020cracku} having pixel-level annotations. Note that many other pavement damage datasets only have the image- and block-level annotations, such as the ones in Table \ref{tab:existing_datasets}. However, this study restricts this part of the analysis to those with a focus on damage segmentation because any pixel-level annotations can be easily converted into the image- and block-level annotations but the image- and block-level annotations cannot be transformed into the pixel-level ones.
    
\begin{figure}[htbp]
    	\centering
    	\subfloat[\label{tab:dataset_compare}]{
    		\begin{minipage}[t]{0.5\textwidth}
    			\centering
    			\includegraphics[width=\textwidth]{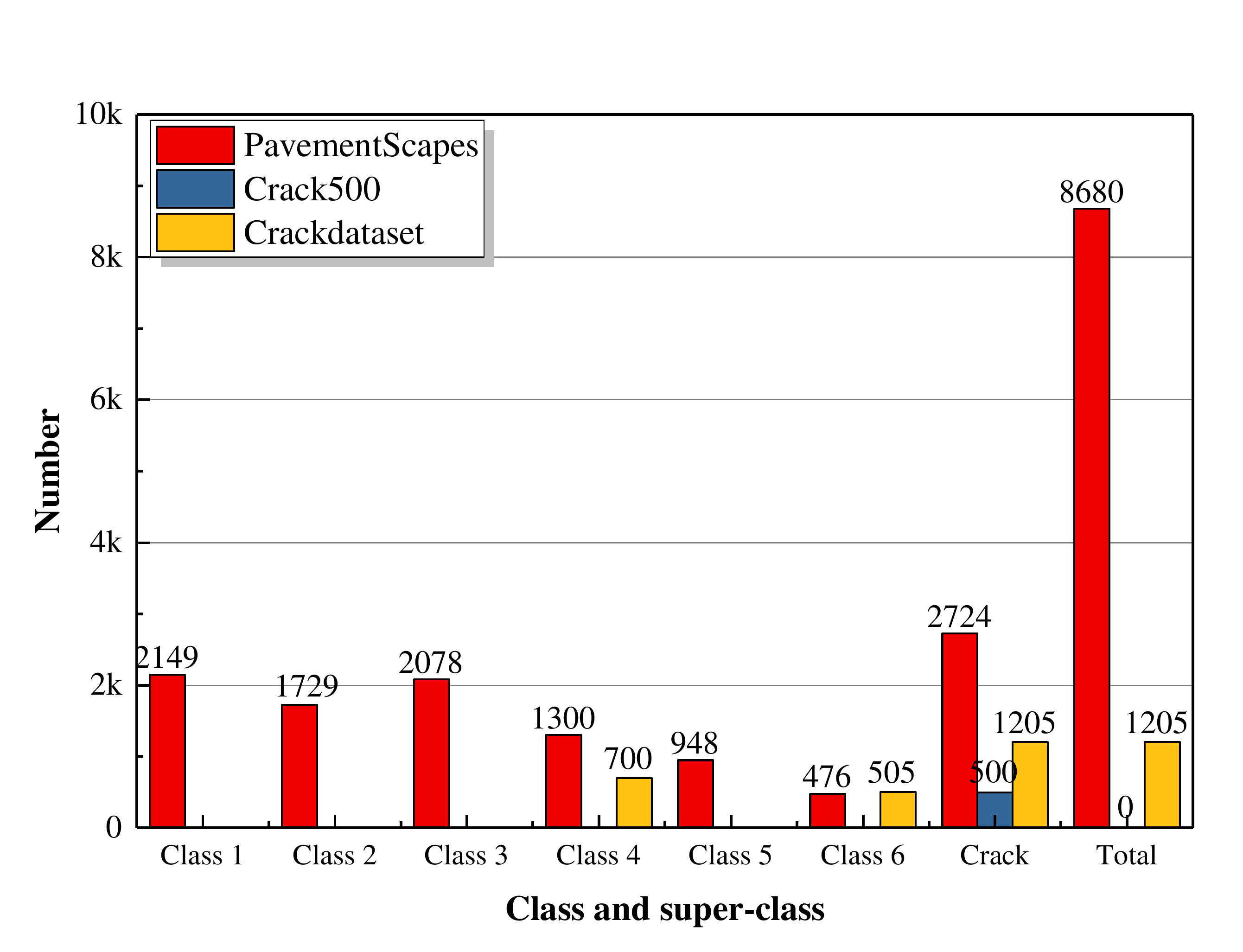}
    		\end{minipage}
    	}
    	\subfloat[\label{tab:train_val_test}]{
    		\begin{minipage}[t]{0.5\textwidth}
    			\centering
    			\includegraphics[width=\textwidth]{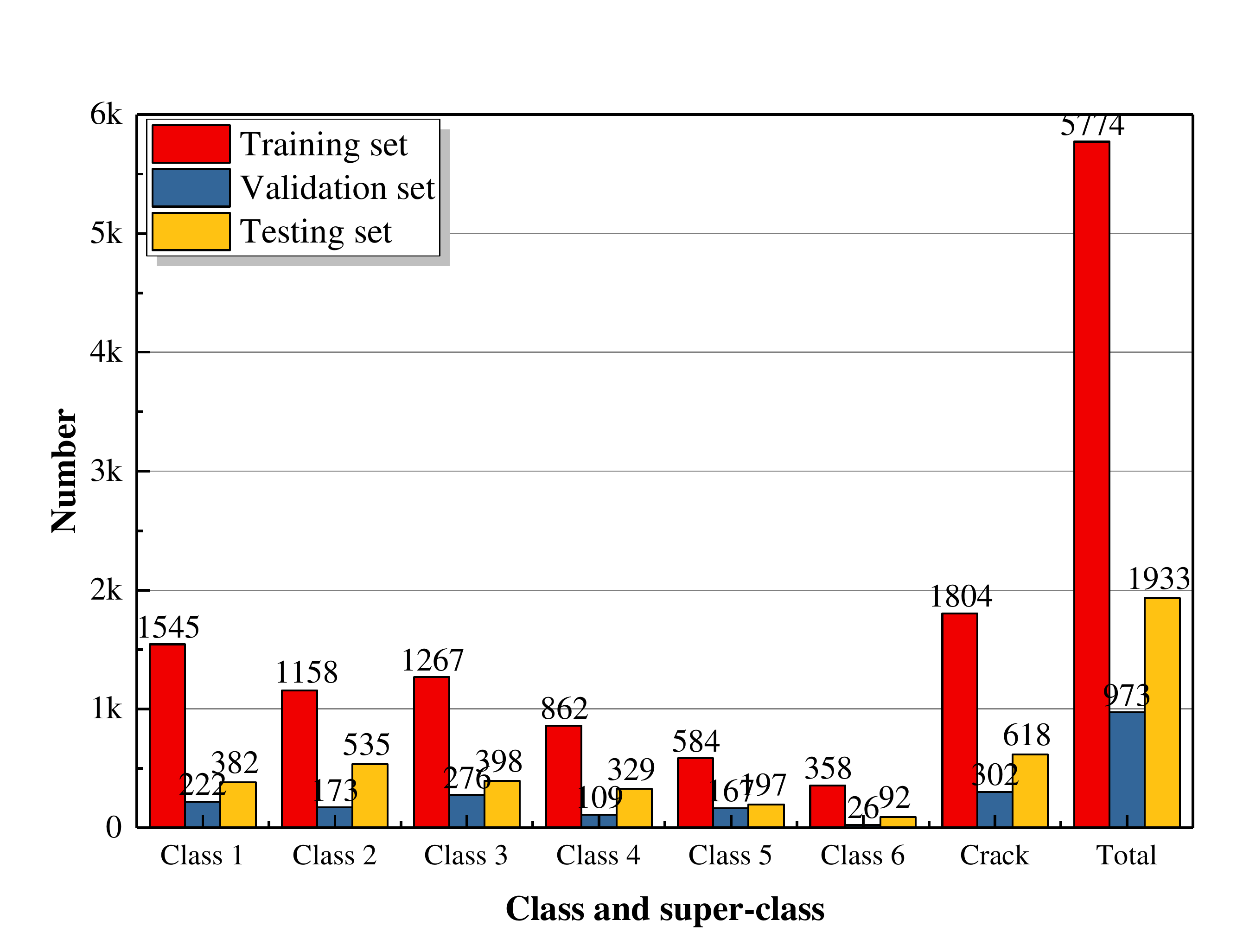}
    		\end{minipage}
    	}\\
    	\caption{Damage distribution: (a) comparison of different datasets and (b) distribution in the training, validation, and testing sets of the Pavementscapes dataset. Class 1, Class 2, Class 3, Class 4, Class 5, and Class 6 stand for ``pothole'', ``rut'', ``repair area'', ``longitudinal crack'', ``lateral crack'', and ``alligator crack'', respectively. The number of the ``crack'' super-class are the sum of the numbers of Class 4, Class 5, and Class 6.}\label{fig:damage_distribution}
    \end{figure}

    \paragraph{Damage distribution.} Figure \ref{tab:dataset_compare} presents the numbers of damages across individual classes on the Pavementscapes dataset. In terms of the overall composition, the distribution of different damages is not uniform but close to the distribution of these damages in the real world, which makes deep neural networks easy to learn different damage features. Figure \ref{tab:train_val_test} presents the damage distributions in the training, validation, and testing sets. The distributions on the three sets are similar to the distribution of the proposed dataset. The ratio of the damage numbers in the training, validation, and testing sets is about 6:1:2.
    
    Figure \ref{tab:dataset_compare} also compares the Pavementscapes dataset with the Crack500 and CrackDataset datasets. The proposed dataset exceeds the other two datasets in the inherently different configurations. The Pavementscapes dataset involves different pavement surface damages in wide roads (at least one lane with 3.75 m), whereas the Crack500 and CrackDataset datasets are only composed of pavement crack scenes. As a result, the Pavementscapes dataset exhibits six types of pavement visual damages, while the other two datasets only include crack damages. Besides, the other two datasets do not refine the crack category into some sub-categories, such as longitudinal, lateral, and alligator cracks.
    
\begin{figure}[htbp]
    	\centering
    	\subfloat[\label{tab:service_year}]{
    		\begin{minipage}[t]{0.5\textwidth}
    			\centering
    			\includegraphics[width=\textwidth]{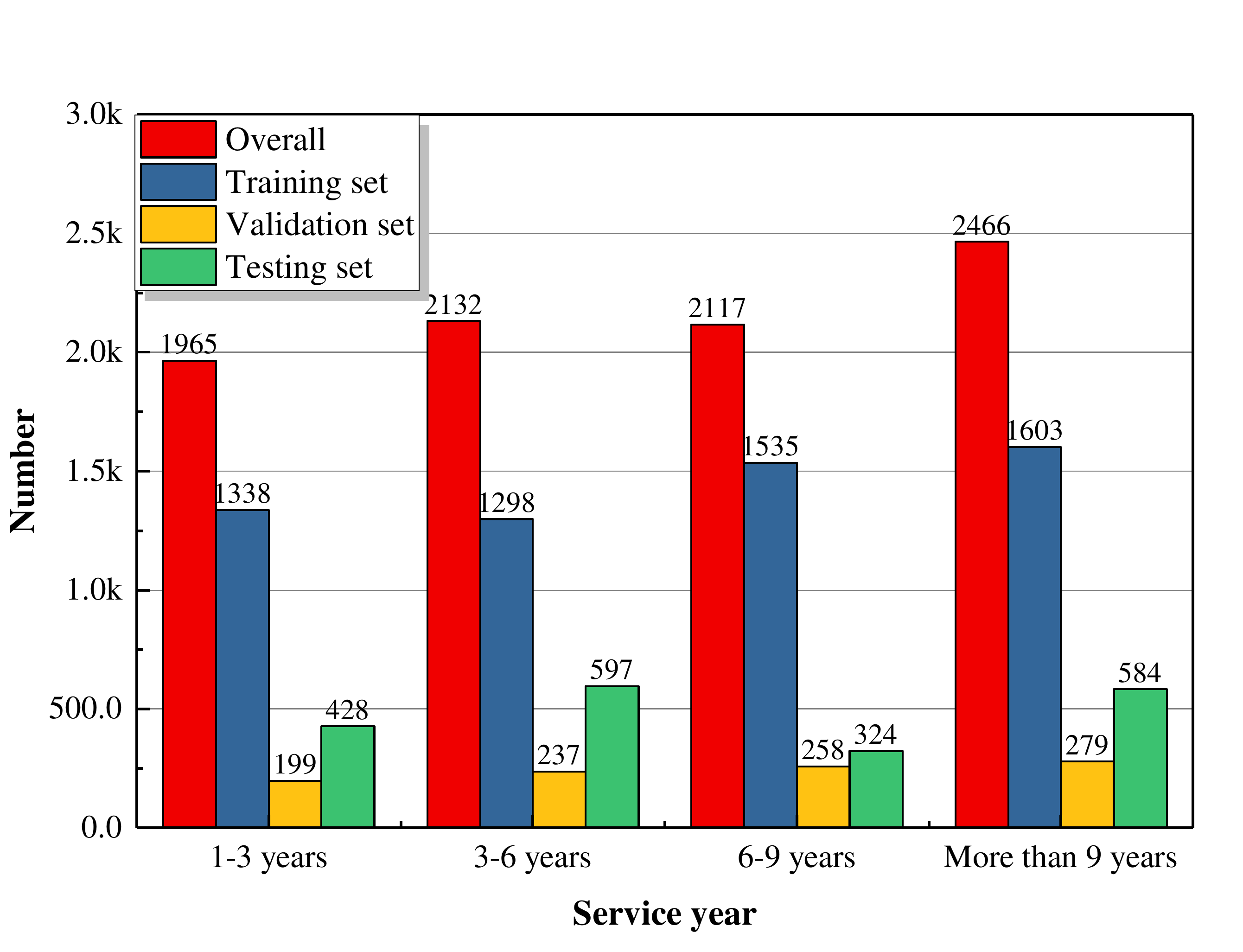}
    		\end{minipage}
    	}
    	\subfloat[\label{tab:weather}]{
    		\begin{minipage}[t]{0.5\textwidth}
    			\centering
    			\includegraphics[width=\textwidth]{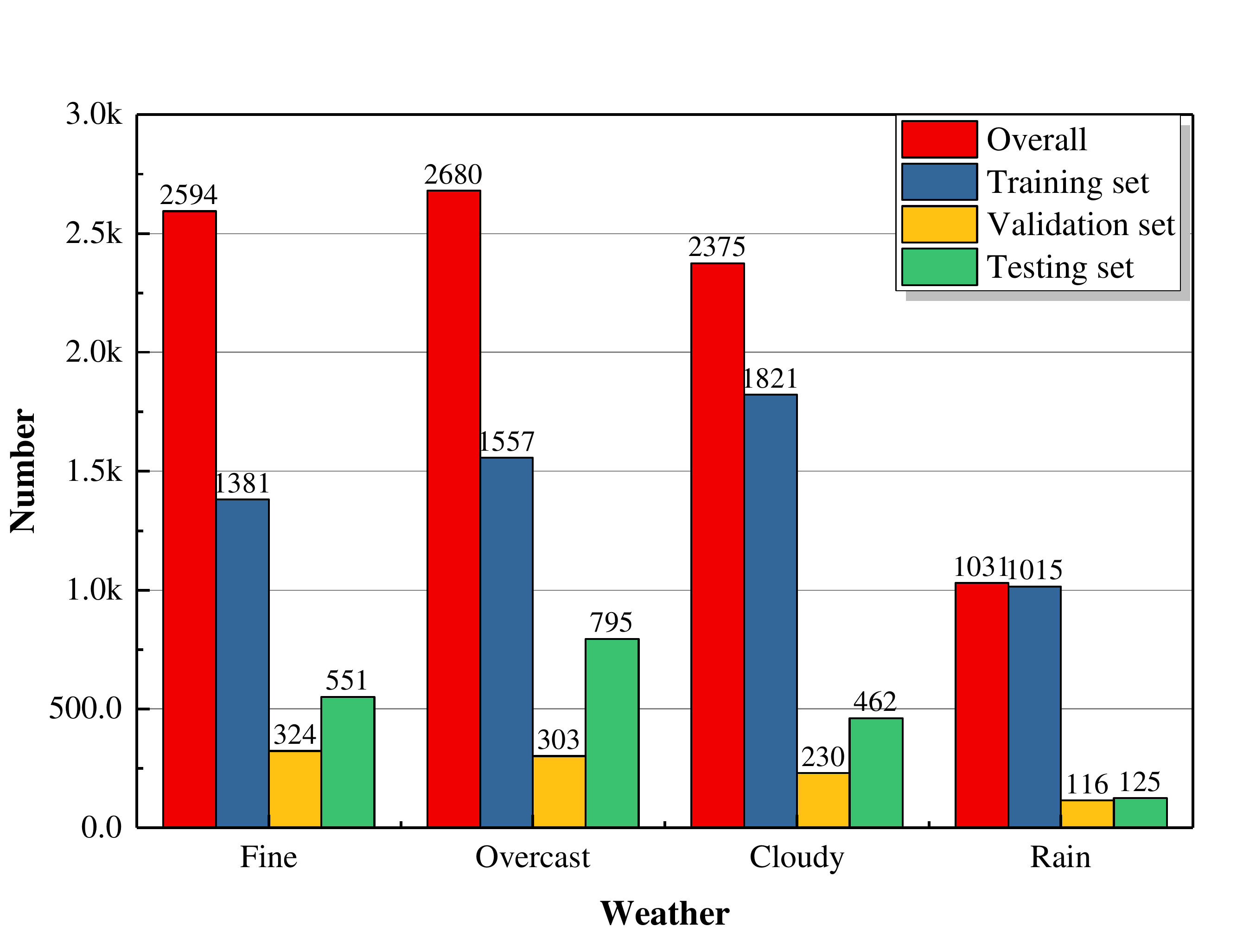}
    		\end{minipage}
    	}\\
    	\subfloat[\label{tab:materials}]{
    		\begin{minipage}[t]{0.5\textwidth}
    			\centering
    			\includegraphics[width=\textwidth]{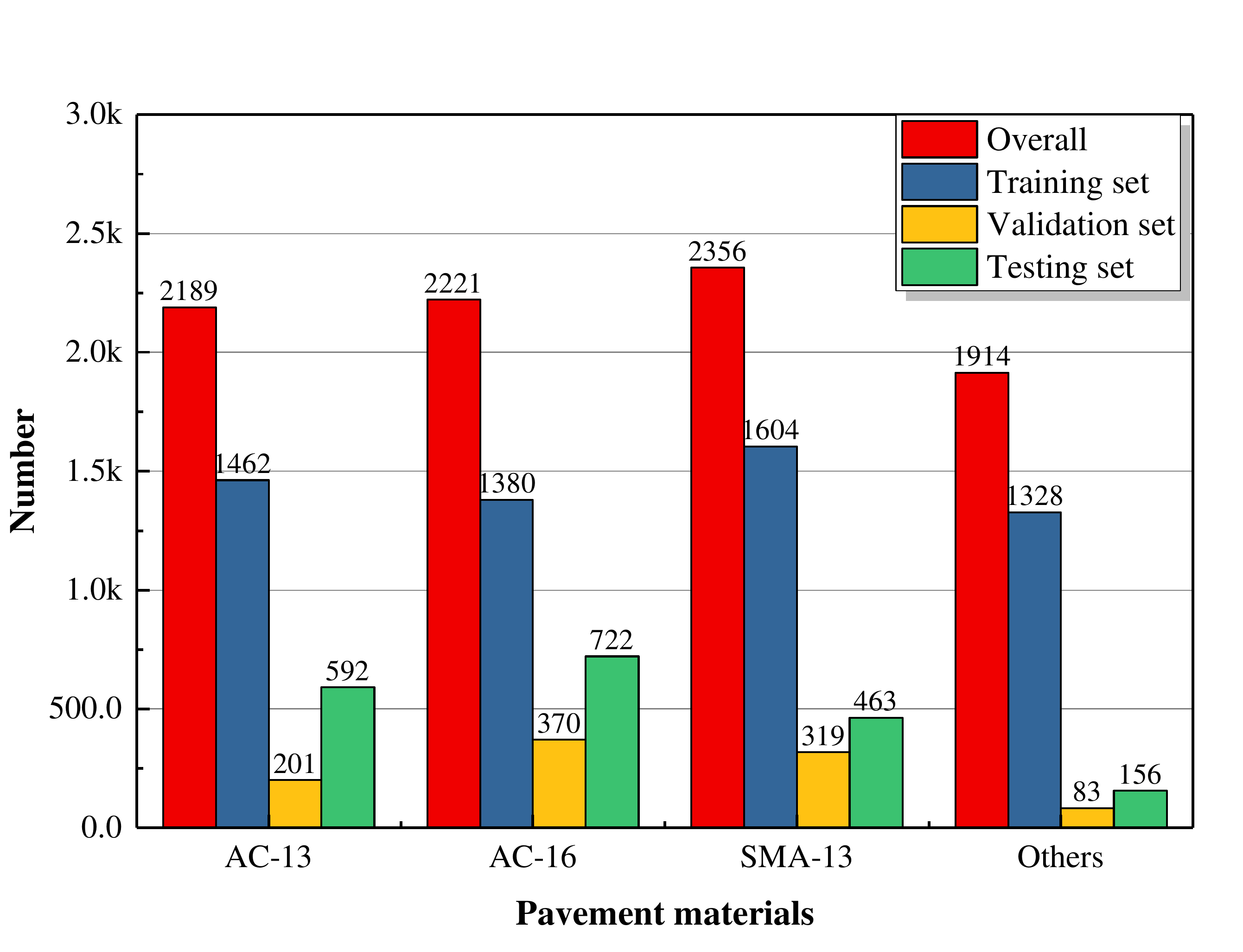}
    		\end{minipage}
    	}\\
    	\caption{Numbers of pavement damages under different real-world conditions: (a) service years, (b) weather when the images were collected, and (c) pavement materials.}\label{fig:scene_complexity}
    \end{figure}

    \paragraph{Scene complexity.} The scene complexity is assessed on the Pavementscapes dataset, where the dataset is split based on the environmental conditions when the images were collected. Figure \ref{fig:scene_complexity} shows the image distributions under various real-world conditions. The distributions of the pavement images are uniform under various service years, sunlight conditions, and surface materials, which ensures the comprehensiveness of the proposed dataset. Once given the proposed dataset, deep neural networks can learn the knowledge of damage features in different environments, which can ensure the generality of deep learning models.
    
    \paragraph{Annotation accuracy.} The quality of the annotations is assessed in the study. First, 50 images were randomly selected and labeled three times by different annotators following the quality control in Section \ref{sec:class}. More than 92\% of pixels were labeled as the same classes. Second, the annotators were required to select a ``background'' label if they did not have the full certainty about the pixel class, such as some small damage instances. After excluding the ``background'' pixels, we recounted 95\% agreement in the annotations of the 50 images. Finally, all annotations of different categories of cracks were coarsely annotated as ``crack'' super-category, for example, the longitudinal-crack pixels are annotated as ``crack''. In the 50 images, 98\% pixels in the coarse annotations were defined with the same category. Therefore, the Pavementscapes dataset has annotations with good and stable quality.
    
    \paragraph{Non-iconic views.} One goal of the proposed dataset is to collect non-iconic pavement images. Most pavement image datasets only include the images on an iconic view, such as Crack500 and CrackDataset. Besides, the current deep learning systems perform fairly well on iconic views. However, in many pavement inspection projects, many pavement images present many damages on non-iconic views, such as partially occluded cracks. Unfortunately, the current deep learning systems struggle to segment objects on the non-iconic views. The Pavementscapes dataset collects 4,828 damage instances on an iconic view (e.g., repair areas in the third row and left column of Figure \ref{fig:examples}) and 3,852 damages on different non-iconic views  (e.g., alligator cracks in the second row and right column of Figure \ref{fig:examples}). The proposed dataset with both iconic and non-iconic views allows to train deep neural networks with reasonable segmentation performance on different views.
 
\section{Experiment}
\label{sec:experiments}

This section provides the numerical experiment that uses the Pavementscapes dataset to train and test top-performing deep neural networks. Sections \ref{sec:tasks_metrics} and \ref{sec:implementation} introduces the metrics and implementation details in the experiment, respectively. Section \ref{sec:r_segmentation} discusses the performances of convolution- and attention-based deep neural networks on pavement damage segmentation. Finally, Section \ref{sec:recommendationss} provides the recommendations and future scopes on the damage segmentation task.

    \subsection{Metrics}
    \label{sec:tasks_metrics}
	
	This experiment uses five metrics to evaluate the performance of deep neural networks for damage segmentation: pixel accuracy (PA), mean intersection over union (mIoU), and expected calibration error (ECE), floating point operations (FLOPs),  and network parameters.
	
	\paragraph{Pixel accuracy.} Let $\Omega=\{\omega_0,\omega_1,\dots,\omega_m\}$ be the set of classes, where $\omega_0$ is the ``background'' class and $\omega_i$ is one of the damage classes, $i=1,\dots.m$. Given an image with $T$ pixels, the \emph{pixel accuracy} is defined as
	\begin{equation}\label{con:pu}
		PA=\frac1{|T|}\sum_{j=1}^{|T|} \one_{\omega(j)}\left(\widehat \omega(j)\right),
	\end{equation}
	where $\omega_{\ast}(j)$ and $\widehat \omega(j)$ are the labeled and predicted class of pixel $j$, and $\one$ is the indicator function of class $\omega(j)$. Note that the pixels belonging to the ``background'' class do not consider in the metric, as do in many benchmark datasets \cite{cordts2015cityscapes,deng2009imagenet,everingham2015pascal,lin2014microsoft}. 
	
	\paragraph{Mean intersection over union.} This metric measures overlap between labeled and predicted areas of a object as
	\begin{equation}\label{con:iou}
			mIoU=\frac{1}{m}\sum_{i=1}^M\frac{|\boldsymbol{G}^i \cap \boldsymbol{P}^i|}{|\boldsymbol{G}^i \cup \boldsymbol{P}^i|}
	\end{equation}
	where $\boldsymbol{G}^i$ and $\boldsymbol{P}^i$ are the ground truth and predicted pixel set of class $i$. This experiment do not consider the IoU of the ``background'' class, as do in many benchmark datasets \cite{cordts2015cityscapes,deng2009imagenet,everingham2015pascal,lin2014microsoft}.
	
	\paragraph{Expected calibration error.} In a learning system, a network should not only make correct predictions but also show when it may fail. The \emph{confidence} of a network is defined as a mass of belief supporting the hypothesis that the prediction of a network is correct. This experiment utilizes the \emph{expected calibration error} (ECE) \cite{guo2017calibration} to measure the confidence of a network and calibrate whether its confidence matches its accuracy. First, the \emph{prediction confidence} of pixel $j$ is defined as  
	\begin{equation}\label{con:pconfidence}
		pc(j)=\widehat{p}(\omega_{\ast}(j)),
	\end{equation}
	where $\widehat{p}(\omega_{\ast}(j))$ is the predicted probability for pixel $j$ in its true class. Let $b_k$ be the bin of pixels whose prediction confidence falls into the interval $(\frac{k-1}{K}, \frac{k}{K}]$, $k=1,\dots,K$. The accuracy and confidence of bin $b_k$ are then computed, respectively, as
	\begin{subequations}\label{con:acbin}
		\begin{equation}\label{con:abin}
			ac(b_k)=\frac{1}{|b_k|}\sum_{j \in b_k} \one_{\omega(j)}\left(\widehat \omega(j)\right),
		\end{equation}
		\begin{equation}\label{con:cbin}
			co(b_k)=\frac{1}{|b_k|} \sum_{j \in b_k} pc(j).
		\end{equation}
	\end{subequations}
	A network is well calibrated with $ac(b_k)\approx co(b_k)$ for all bins, and the ECE\footnote{The code of ECE is available at \url{https://github.com/tongzheng1992/E-FCN}, which has been released by the first author in the previous study  \cite{tong2021evidentialfcn}.} is defined as
	\begin{equation}\label{con:sum_ece}
		ECE=\frac{\sum_{k=1}^K |b_k| \times |co(b_k)-ac(b_k)|}{\sum_{k'=1}^K |b_k'|}.
	\end{equation}
	The ECE in this experiment does not consider the ``background'' pixels.

	\paragraph{Floating point operations and network parameters.} This experiment measures the complexity of a deep neural network using floating point operations (FLOPs) and network parameters. FLOPs are widely used to describe how many operations are required to run a single instance in a deep neural network \cite{dosovitskiy2020image,guo2021cmt,xie2017aggregated}; calculation processes can be found in \cite{hunger2005floating}. Lower values of FLOPs and network parameters always mean that an algorithm processes a new instance with fewer computation costs.
	
	\subsection{Implementation details}
	\label{sec:implementation}
	
	The experiment only focuses on deep neural networks for pavement damage segmentation because many previous studies \cite{cha2017deep,cha2018autonomous,tong2020advances,tong2017recognition} have demonstrated that deep learning completely outperforms the other machine learning algorithms involving manual feature engineerings, such as support vector machine and random forest.
	
	The experiment considers the convolution- and attention-based deep neural networks for pavement damage segmentation. For the convolution-based deep neural networks, the Pavementscapes dataset is used to train and test a series of the original FCN models (FCN-32s, FCN-16s, and FCN-8s) \cite{long2015fully}, U-net \cite{ronneberger2015u}, and DeepLabv3+ \cite{chen2018encoder}. These models use the same backbone, VGG16, as shown in Table \ref{tab:vgg}. For the attention-based deep neural networks, four models was considered, including self-attention net \cite{vaswani2017attention}, criss-cross attention (CC-attention) net \cite{huang2019ccnet}, double-attention net \cite{chen20182}, and segmentation transformer \cite{yuan2019segmentation}. The patch size of the attention-based models are $32 \times 32$. Other detailed hyper-parameters of these networks is the same as their original works.

	\begin{table}[]
		\centering
		\caption{Architecture of VGG16 network.}\label{tab:vgg}
		\begin{tabular}{ccc}
			\hline
			Stage                    & Layer    & Details                           \\ \hline
			\multirow{3}{*}{Stage 1} & Conv 1-1 & 3$\times$3 Conv. 16 $ReLu$ with 1 strides  \\
			& Conv 1-2 & 3$\times$3 Conv. 16 $ReLu$ with 1 strides  \\
			& Pooling  & 2$\times$2 max-pooling with 2 strides    \\ \hline
			\multirow{3}{*}{Stage 2} & Conv 2-1 & 3$\times$3 Conv. 32 $ReLu$ with 1 strides  \\
			& Conv 2-2 & 3$\times$3 Conv. 32 $ReLu$ with 1 strides  \\
			& Pooling  & 2$\times$2 max-pooling with 2 strides    \\ \hline
			\multirow{4}{*}{Stage 3} & Conv 3-1 & 3$\times$3 Conv. 64 $ReLu$ with 1 strides  \\
			& Conv 3-2 & 3$\times$3 Conv. 64 $ReLu$ with 1 strides  \\
			& Conv 3-3 & 3$\times$3 Conv. 64 $ReLu$ with 1 strides  \\
			& Pooling  & 2$\times$2 max-pooling with 2 strides    \\ \hline
			\multirow{4}{*}{Stage 4} & Conv 4-1 & 3$\times$3 Conv. 128 ReLu with 1 strides \\
			& Conv 4-2 & 3$\times$3 Conv. 128 $ReLu$ with 1 strides \\
			& Conv 4-3 & 3$\times$3 Conv. 128 $ReLu$ with 1 strides \\
			& Pooling  & 2$\times$2 max-pooling with 2 strides    \\ \hline
			\multirow{4}{*}{Stage 5} & Conv 5-1 & 3 3 Conv. 256 ReLu with 1 strides \\
			& Conv 5-2 & 3$\times$3 Conv. 256 $ReLu$ with 1 strides \\
			& Conv 5-3 & 3$\times$3 Conv. 256 $ReLu$ with 1 strides \\
			& Pooling  & 2$\times$2 max-pooling with 2 strides    \\ \hline
		\end{tabular}
	\end{table}
	
	During training, all networks use the generalized dice loss function \cite{sudre2017generalised}, which reduces the negative effects of  unbalanced learning set. In this study, the unbalanced learning set means that the number of pixels belonging to different classes are very different, such that the proposed dataset includes a very small number of crack pixels and a very huge number of background pixels in the training set of the Pavementscapes dataset. The phenomenon cannot be avoided because cracks only occupy very small areas in a pavement. Given a pixel with one-hot label $\boldsymbol{y}$ and predicted probabilities $\widehat{\boldsymbol{p}}$, the generalized dice loss function is defined as
	\begin{equation}\label{con:loss}
		\mathcal{L}(\boldsymbol{y}, \widehat{\boldsymbol{p}})= 1- \frac{2\boldsymbol{y}\widehat{\boldsymbol{p}}+1}{\boldsymbol{y}+\widehat{\boldsymbol{p}}+1}.
	\end{equation}
	
	All models are achieved based on TensorFlow 2.8 version. The input image size are set as $1024 \times 2048 \times 1$. The training batch size is 24 and the popular ADAM optimizer with momentum 0.9 and weight decay 1e-4 is used to optimize the networks for backpropagation. Note that some types of data augmentations cannot be used in the Pavementscapes dataset since flips and rotations change the semantics of longitudinal and lateral cracks. The deep neural networks are trained on an Nvidia V100 GPU with 32GB memory.
	
    \subsection{Results of damage segmentation} 
    \label{sec:r_segmentation}
	
    Table \ref{tab:overall_testing} display the overall test performance of the deep neural networks. Some testing examples are shown in \ref{sec:appendix_segmentor}. The attention-based networks exceed the convolution-based ones on PA and mIoU, even though the FLOPs and network parameters of the attention-based models are larger than the ones of the convolution-based ones. In detail, the segmentation transformer model achieves the best segmentation performance, followed by double-attention and CC-attention nets. This demonstrates that the attention-based models outperform the convolution-based ones on pavement damage segmentation. This behavior can be explained by Table \ref{tab:iou_class}. The two types of deep neural networks have similar and high PAs and mIoUs in the ``rut'', ``repair area'', and ``pothole'' classes, but the PAs and mIoUs of convolution-based models in the three crack classes are lower than these of the attention-based models. This is mainly because the attention mechanism allows the attention-based models to focus on the features of some small target objects \cite{niu2021review}, such as the cracks with thin widths. However, the convolution-based feature extraction in the convolution-based models easily ignores the small features, e.g., the boundary areas of cracks and background. Therefore, attention-based deep neural networks have the powerful potential for the accuracy improvement of pavement damage segmentation, especially for some small damages.
    
    \begin{table}[]
    	\centering
    	\caption{Testing performances of deep neural networks on the Pavementscapes dataset. GFLOPs stands for $10^9$ (giga) floating point operations and $M$ means million. The best and second results in each term are marked in bold and italics.}\label{tab:overall_testing}
    	\begin{tabular}{lccccc}
    		\hline
    		& PA/\% & mIoU   & ECE/\% & GFLOPs & Parameter/M \\ \hline
    		FCN-32s \cite{long2015fully} & 66.53 & 51.94 & 22.48  & -       & -           \\
    		FCN-16s \cite{long2015fully} & 67.02 & 52.21 & 22.41  & -       & -           \\
    		FCN-8s  \cite{long2015fully} & 67.32 & 52.98 & 22.30  & \textbf{177}     & 134.3       \\
    		U-net  \cite{ronneberger2015u} & 69.56 & 54.71 & 22.14  & \textit{194}     & 19.4        \\
    		DeepLabv3+ \cite{chen2018encoder} & 71.90 & 57.51 & 21.81  & 783     & 41.1        \\ \hline
    		Self-attention  net \cite{vaswani2017attention} & 73.07 & 58.74 & 21.32  & 619     & \textit{10.5}        \\
    		CC-attention net \cite{huang2019ccnet} & 73.15 & 58.52 & 21.14  & 804     & 10.6        \\
    		Double-attention net \cite{chen20182} & \textit{74.01} & \textit{59.23} & \textit{21.10}  & 338     & \textbf{10.2}        \\
    		Segmentation Transformer \cite{yuan2019segmentation} & \textbf{74.50} & \textbf{59.74} & \textbf{20.95}  & 340     & 10.5        \\ \hline
    	\end{tabular}
    \end{table}

	\begin{table}[]
		\centering
		\caption{Testing IoU results of different damage classes on the Pavementscapes dataset. The best and second results in each term are marked in bold and italics.}\label{tab:iou_class}
		\resizebox{\textwidth}{!}{
		\begin{tabular}{lcccccc}
			\hline
			& Longitudinal crack & Lateral crack & Alligator crack & Pothole                      & Rut   & Repair area \\ \hline
			FCN-32s                  & 25.82              & 27.17         & 25.38           & 67.17                        & 75.92 & 90.17       \\
			FCN-16s                  & 25.36              & 26.51         & 25.56           & 67.32                        & 76.17 & 92.34       \\
			FCN-8s                   & 25.93              & 27.27         & 26.42           & 68.13 & 77.42 & 92.73       \\
			U-net                    & 26.99              & 30.26         & 29.14           & 69.33                        & 80.10  & 92.45       \\
			DeepLabv3+               & 33.12              & 35.25         & 33.84           & 71.31                        & 79.25 & 92.31       \\ \hline
			Self-attention  net      & \textit{35.62}              & 37.25         & 34.92           & 72.82                        & 78.92 & \textit{92.89}       \\
			CC-attention net         & 35.46              & 37.36         & 34.74           & 72.14                        & 78.82 & 92.58       \\
			Double-attention net     & 35.00                 & \textit{38.29}         & \textit{36.12}           & \textit{73.01}                        & \textit{80.22} & 92.74       \\
			Segmentation Transformer & \textbf{36.20}              & \textbf{39.14}         & \textbf{36.42}           & \textbf{73.42}                        & \textbf{80.42} & \textbf{92.81}       \\ \hline
		\end{tabular}}
	\end{table}
    
    Table \ref{tab:overall_testing} indicates that the two types of deep neural networks can accurately segment the ``rut'', ``repair area'', and ``pothole'' instances but cannot do it well on the three types of cracks. This problem derives from the fact that the proportion of crack pixels in the training set is much lower than the ones of other classes. Unfortunately, the fact cannot be changed in the projects of pavement inspection because the cracks and some other damages only occupy very small parts of a pavement. In detail, the proportion of crack pixels is less than 1\%. Thus, the training set of the Pavementscapes dataset is very unbalanced. Such an unbalanced training set makes deep neural networks tend to classify crack pixels to ``background'' during training since the trend does not introduce a large loss, even though the generated dice loss \ref{con:loss} has been used to reduce the negative effect.  This behavior demonstrates that more advanced loss functions should be considered in the future to train the deep neural networks for the pavement damage segmentation task.
    
\begin{figure}[htbp]
    	\centering
    	\subfloat[\label{tab:pd_fcn32}]{
    		\begin{minipage}[t]{0.32\textwidth}
    			\centering
    			\includegraphics[width=\textwidth]{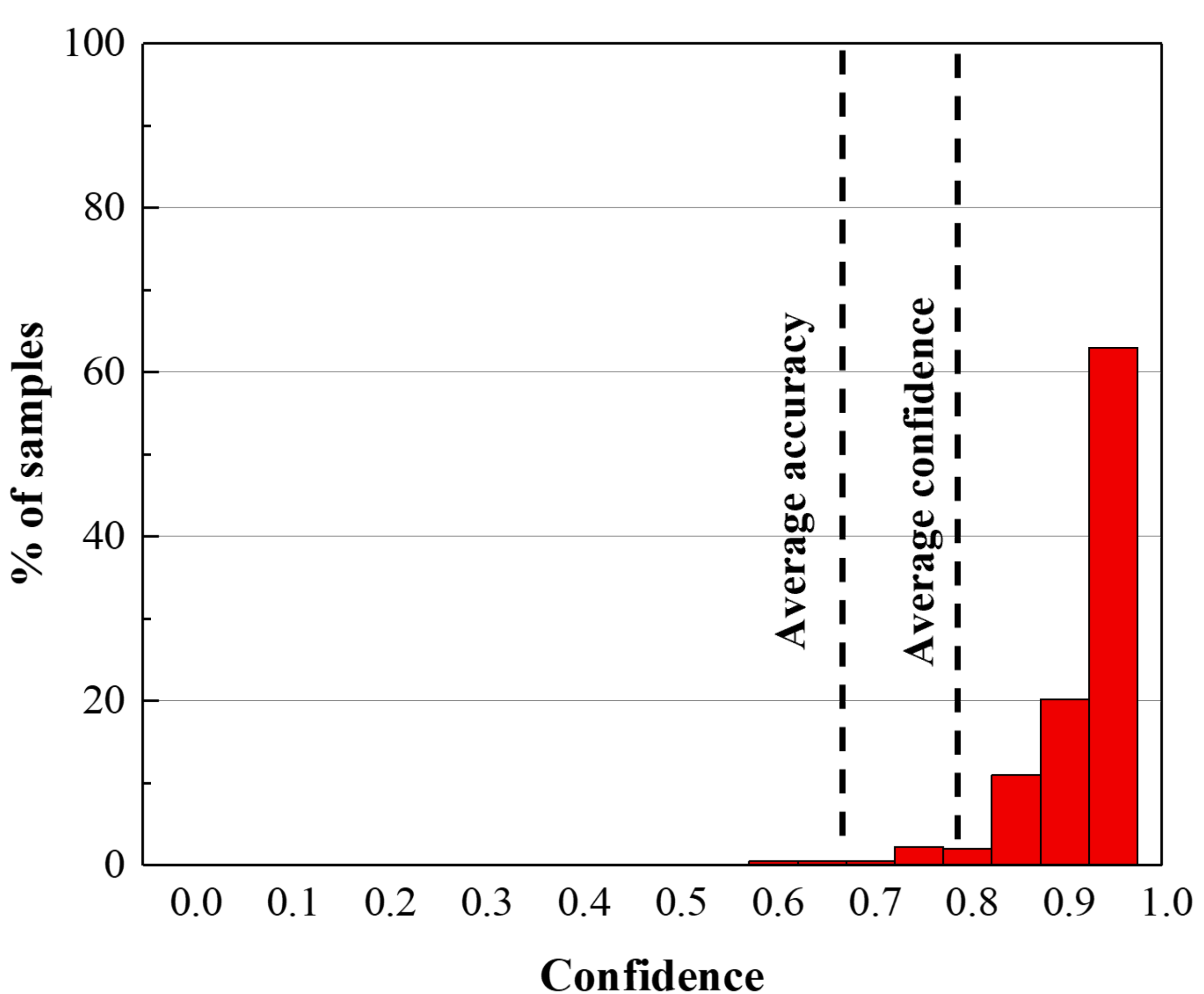}
    		\end{minipage}
    	}
    	\subfloat[\label{tab:pd_fcn16}]{
    		\begin{minipage}[t]{0.32\textwidth}
    			\centering
    			\includegraphics[width=\textwidth]{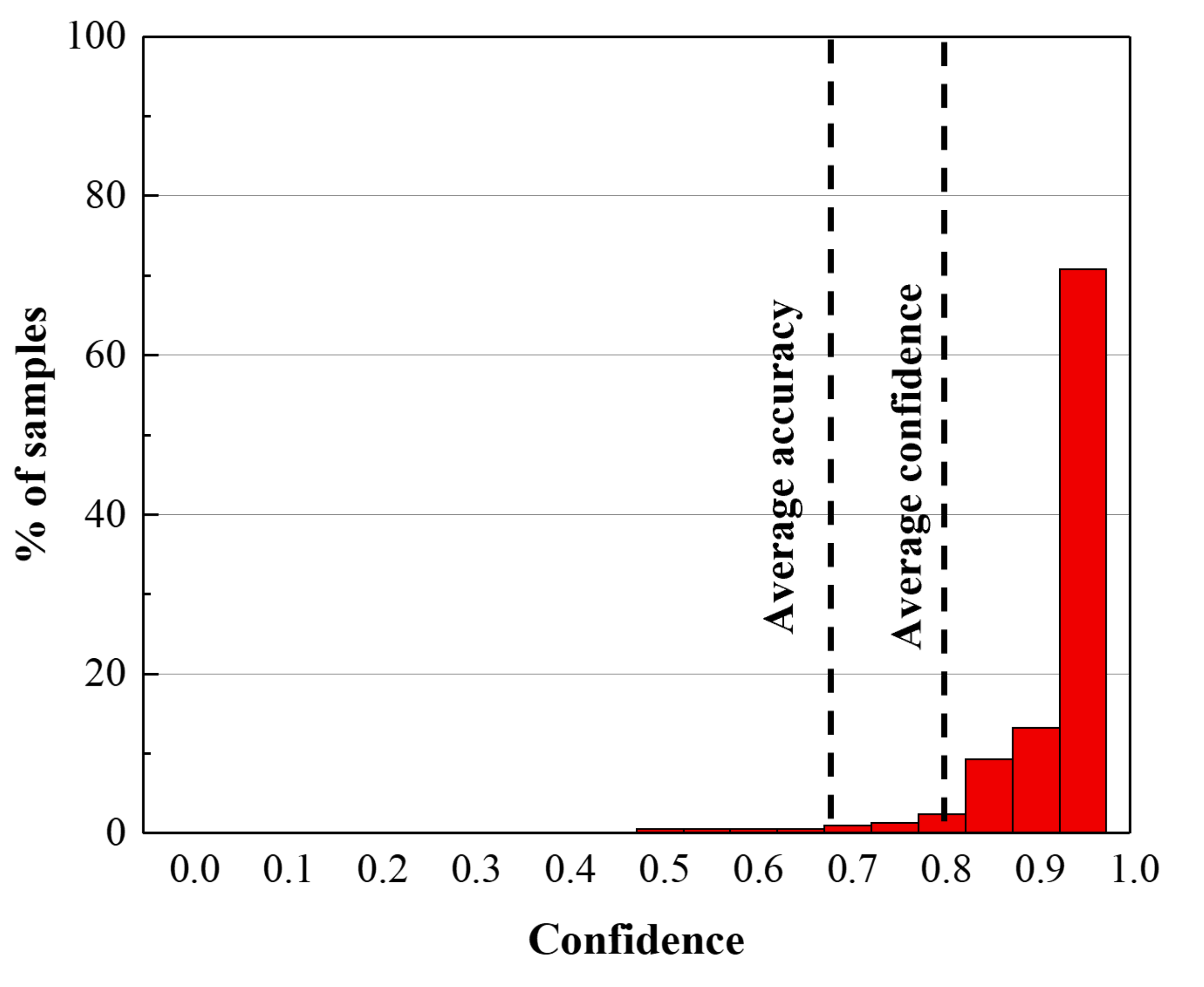}
    		\end{minipage}
    	}
    	\subfloat[\label{tab:pd_fcn8}]{
    		\begin{minipage}[t]{0.32\textwidth}
    			\centering
    			\includegraphics[width=\textwidth]{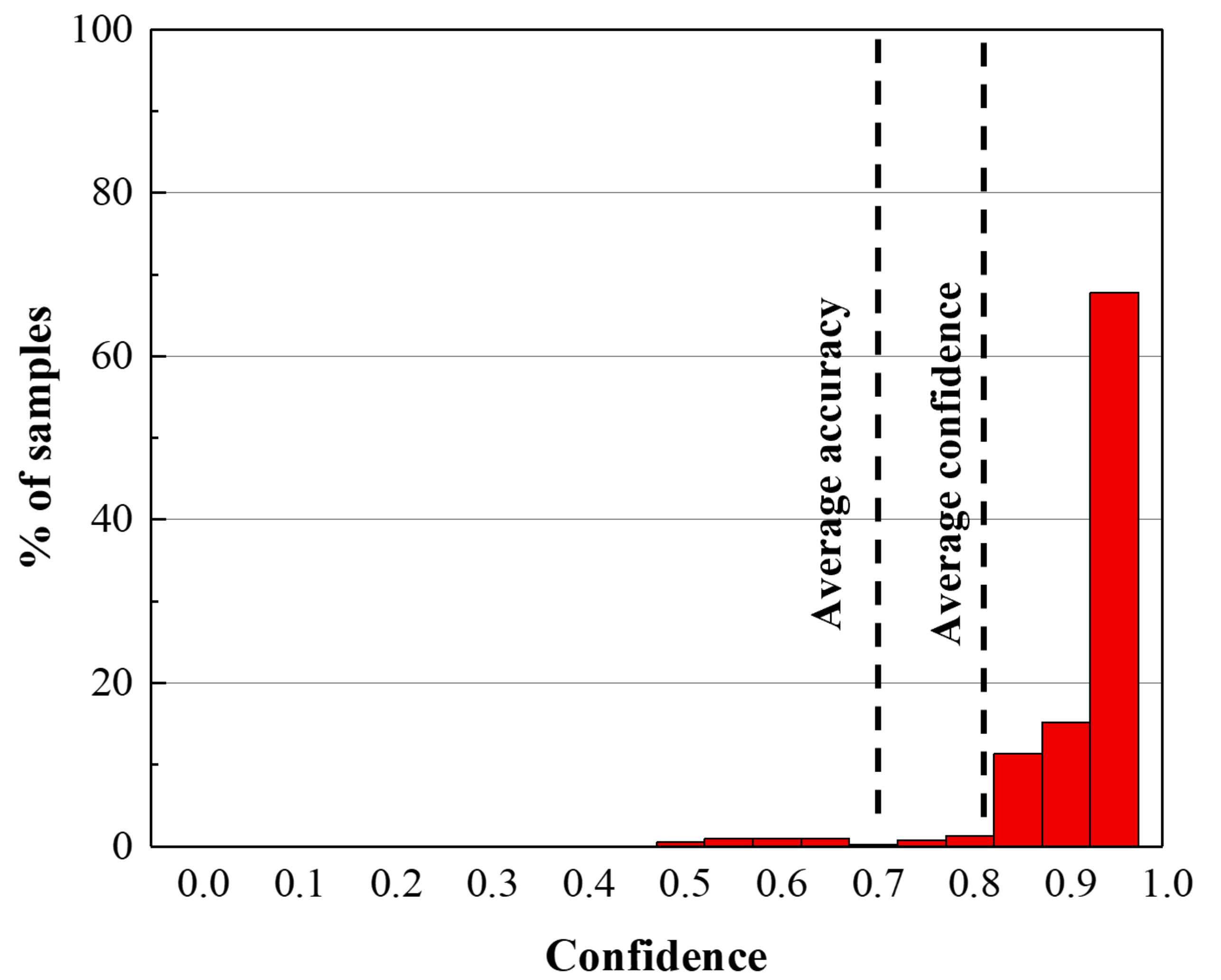}
    		\end{minipage}
    	}\\
    	\subfloat[\label{tab:pd_unet}]{
    		\begin{minipage}[t]{0.32\textwidth}
    			\centering
    			\includegraphics[width=\textwidth]{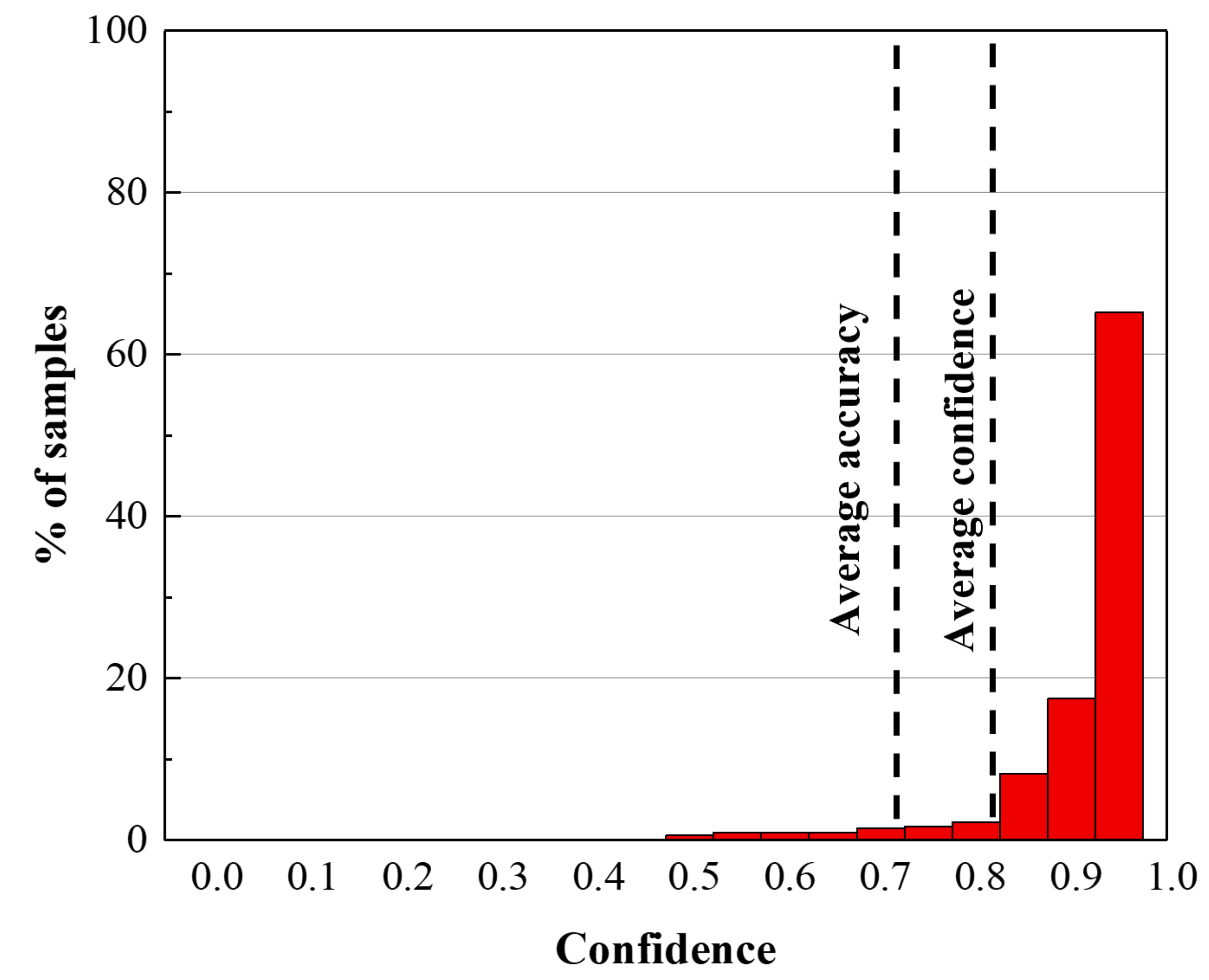}
    		\end{minipage}
    	}
    	\subfloat[\label{tab:pd_deeplab}]{
    		\begin{minipage}[t]{0.32\textwidth}
    			\centering
    			\includegraphics[width=\textwidth]{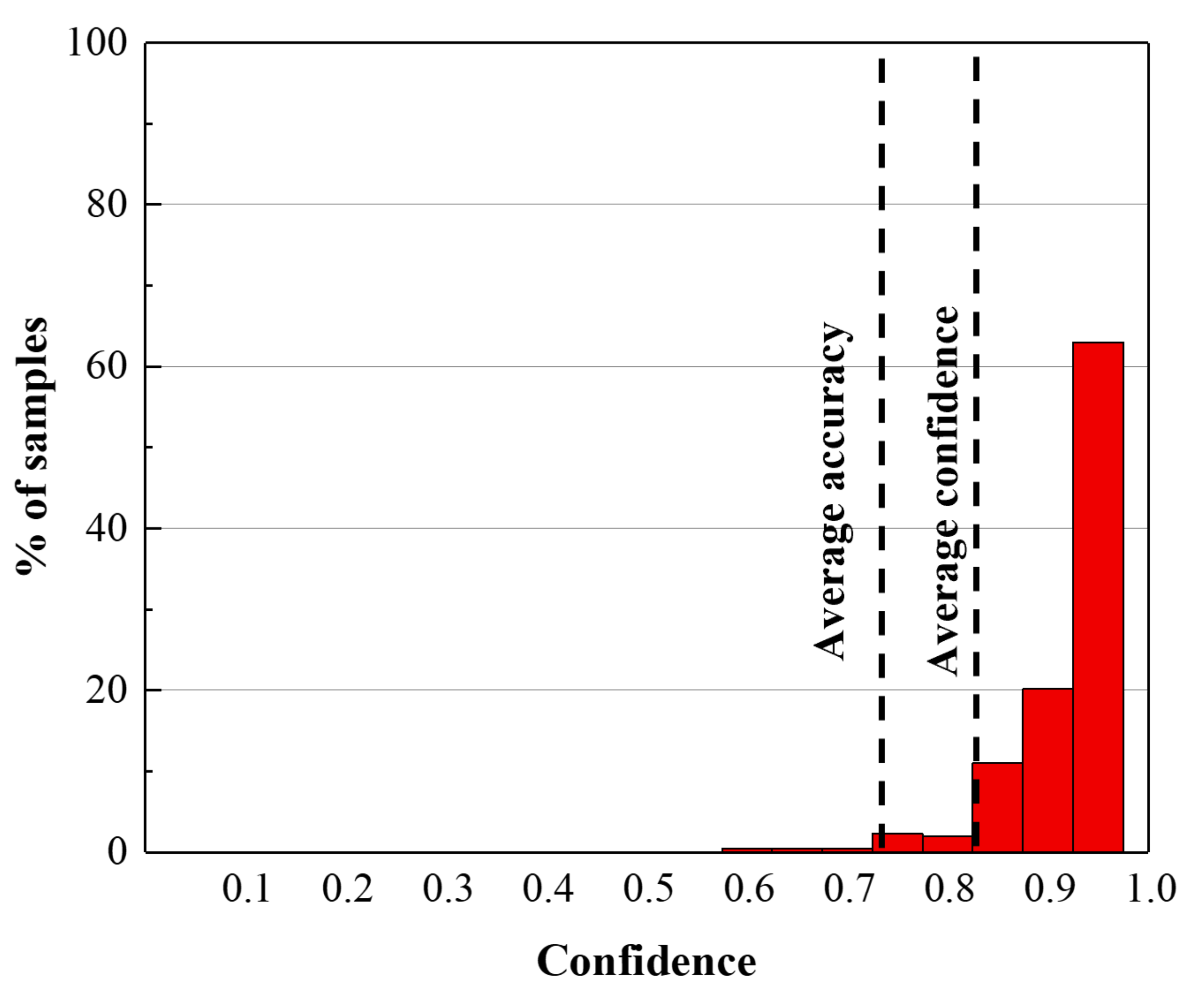}
    		\end{minipage}
    	}
    	\subfloat[\label{tab:pd_self_attention}]{
    		\begin{minipage}[t]{0.32\textwidth}
    			\centering
    			\includegraphics[width=\textwidth]{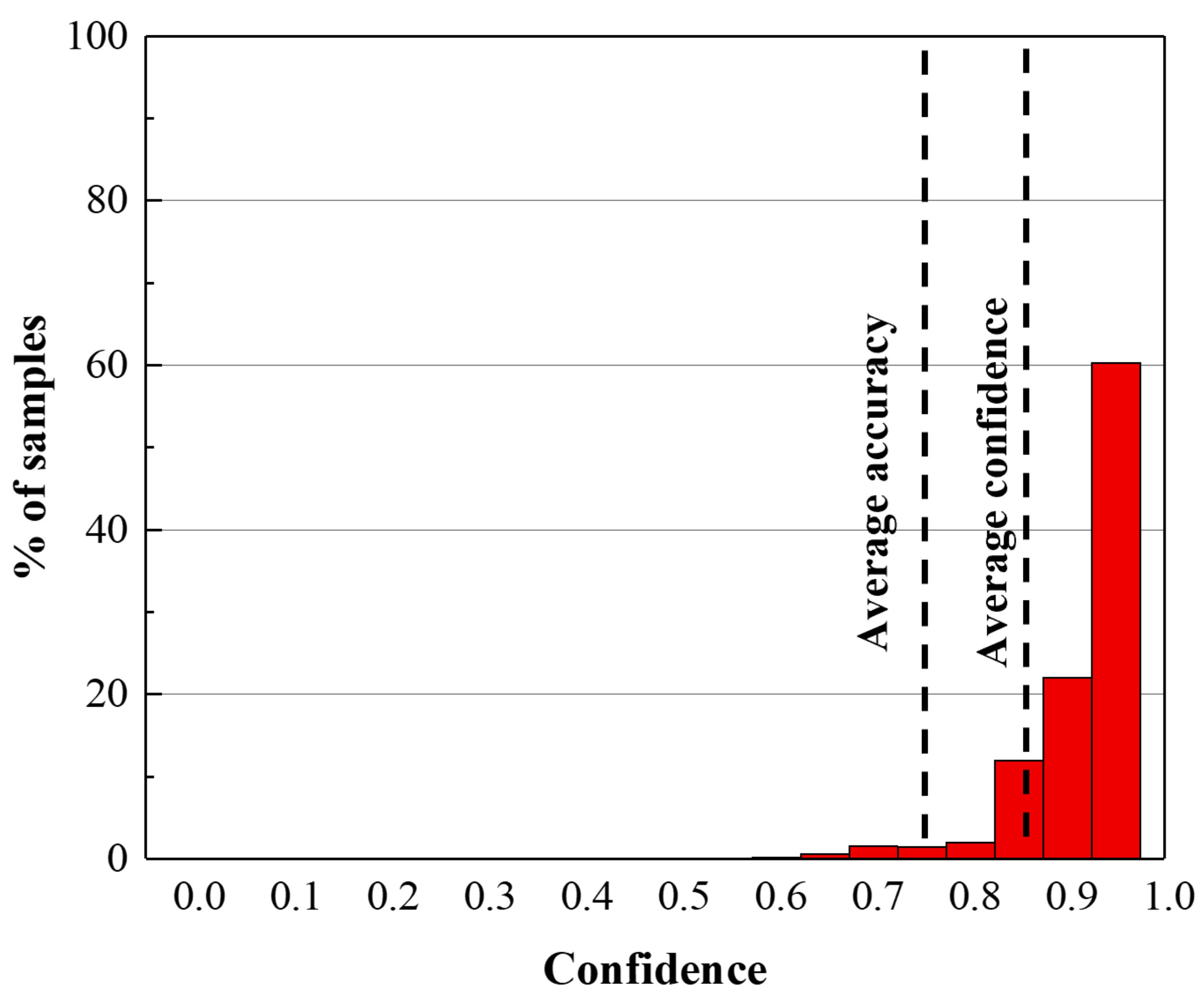}
    		\end{minipage}
    	}\\
    	\subfloat[\label{tab:cc_attention}]{
    		\begin{minipage}[t]{0.32\textwidth}
    			\centering
    			\includegraphics[width=\textwidth]{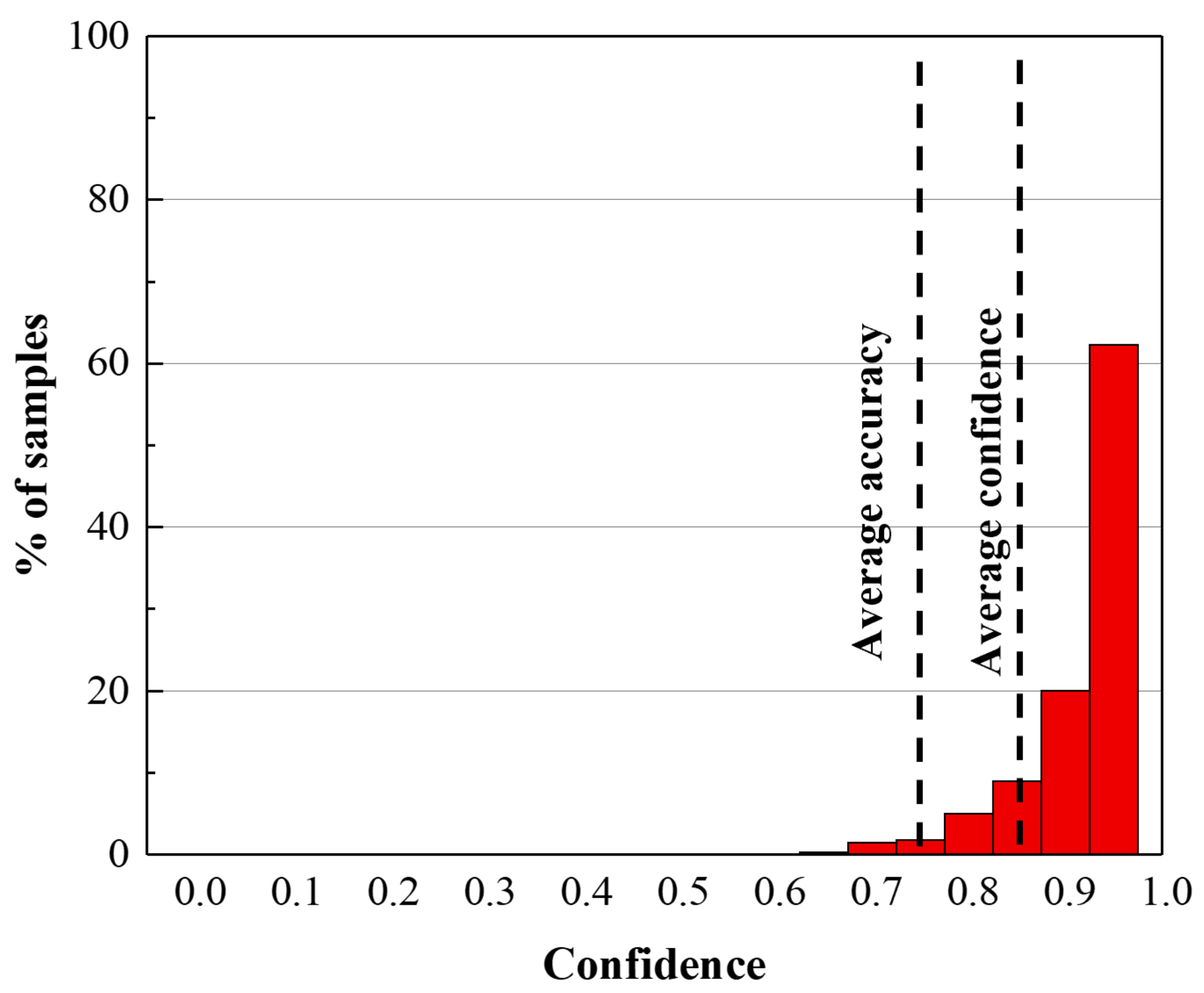}
    		\end{minipage}
    	}
    	\subfloat[\label{tab:pd_double_attention}]{
    		\begin{minipage}[t]{0.32\textwidth}
    			\centering
    			\includegraphics[width=\textwidth]{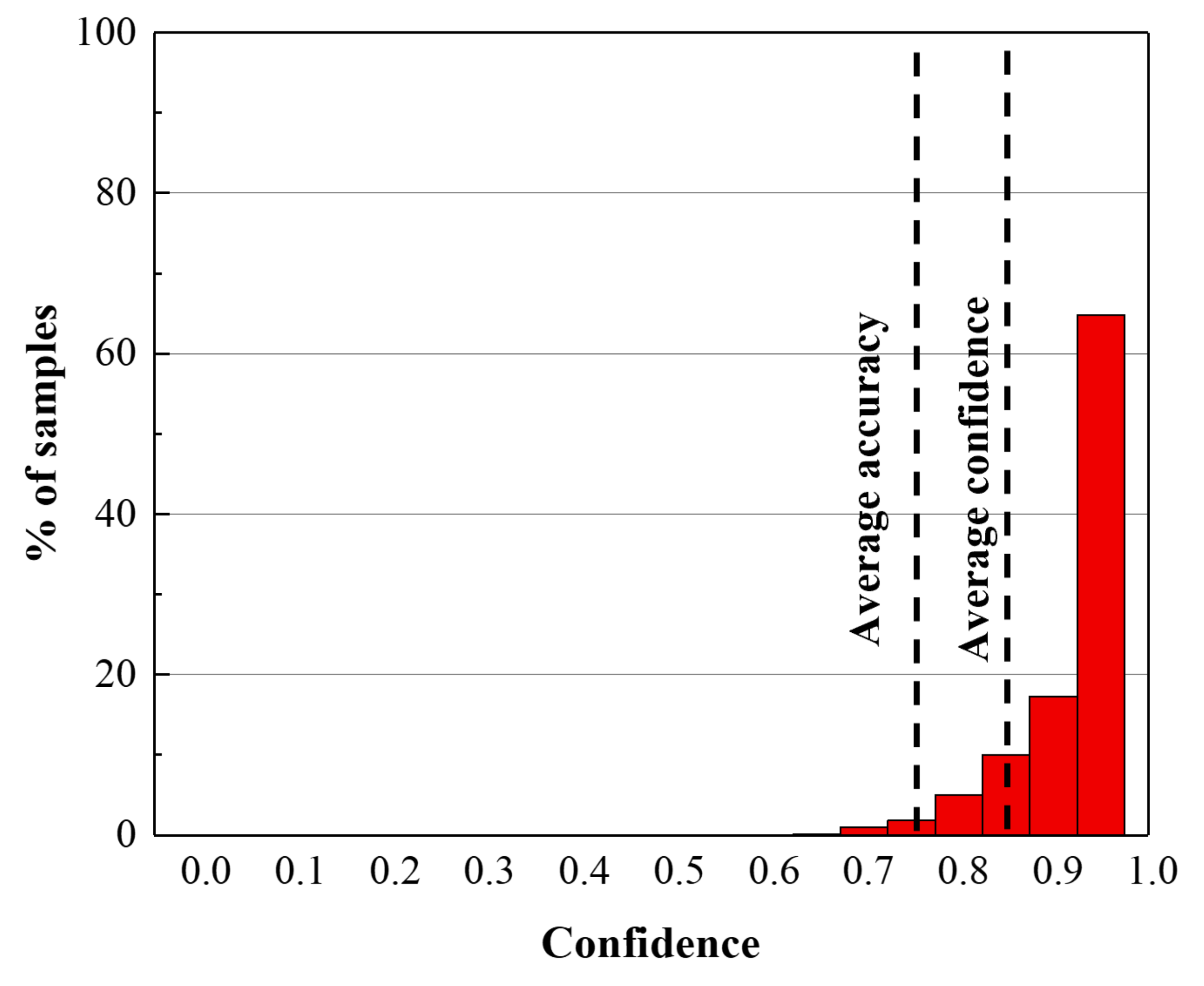}
    		\end{minipage}
    	}
    	\subfloat[\label{tab:pd_transformer}]{
    		\begin{minipage}[t]{0.32\textwidth}
    			\centering
    			\includegraphics[width=\textwidth]{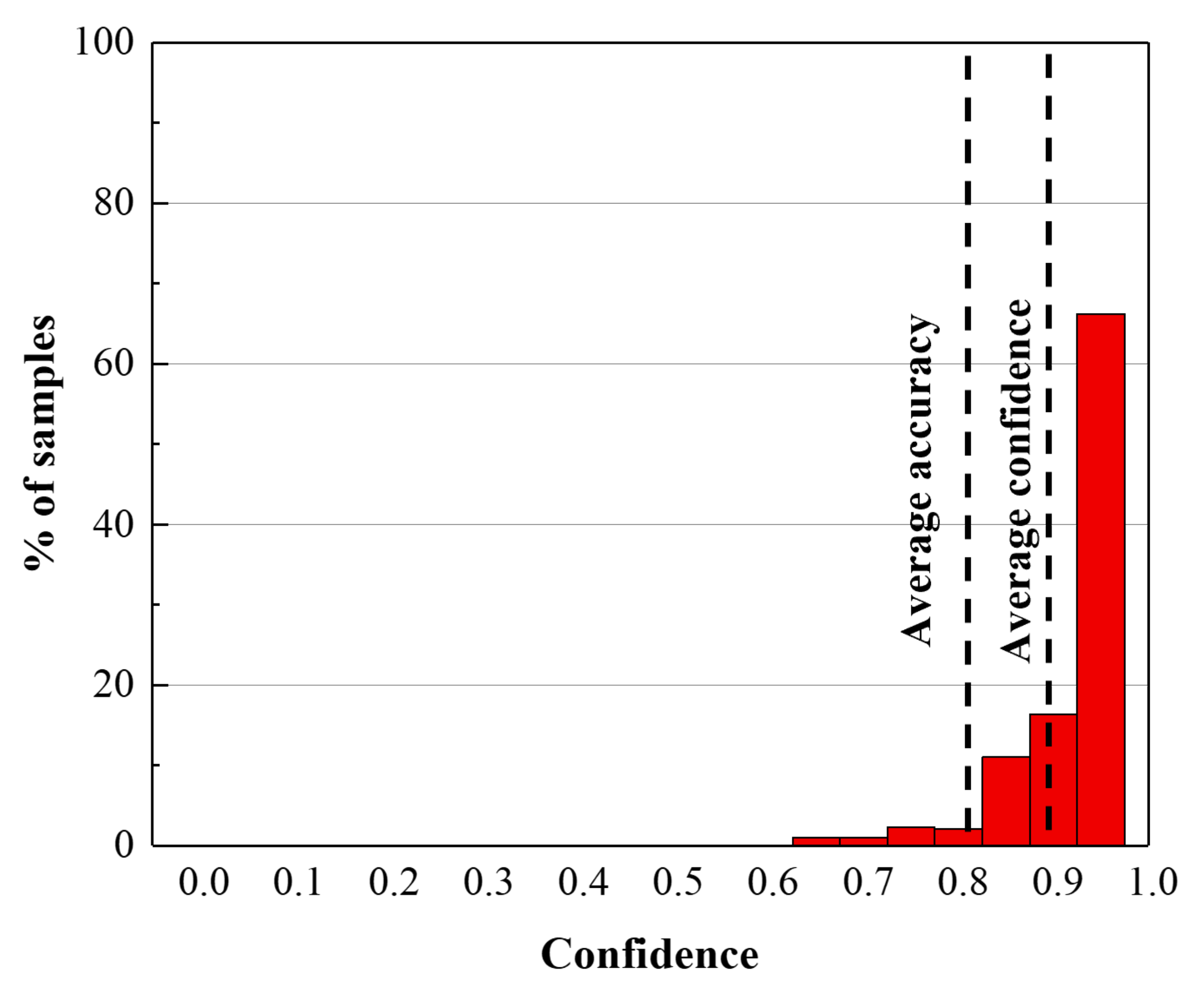}
    		\end{minipage}
    	}\\
    	\caption{Testing pixel distribution on the Pavementscapes dataset: (a) FCN-32s, (b) FCN-16s, (c) FCN-8s, (d) U-net, (e) DeepLabv3+, (f) Self-attention net, (g) CC-attention net, (h) Double-attention net, (i) Segmentation transformer.}\label{fig:pixel_distribution}
    \end{figure}

\begin{figure}[htbp]
		\centering
		\subfloat[\label{tab:ece_fcn32}]{
			\begin{minipage}[t]{0.32\textwidth}
				\centering
				\includegraphics[width=\textwidth]{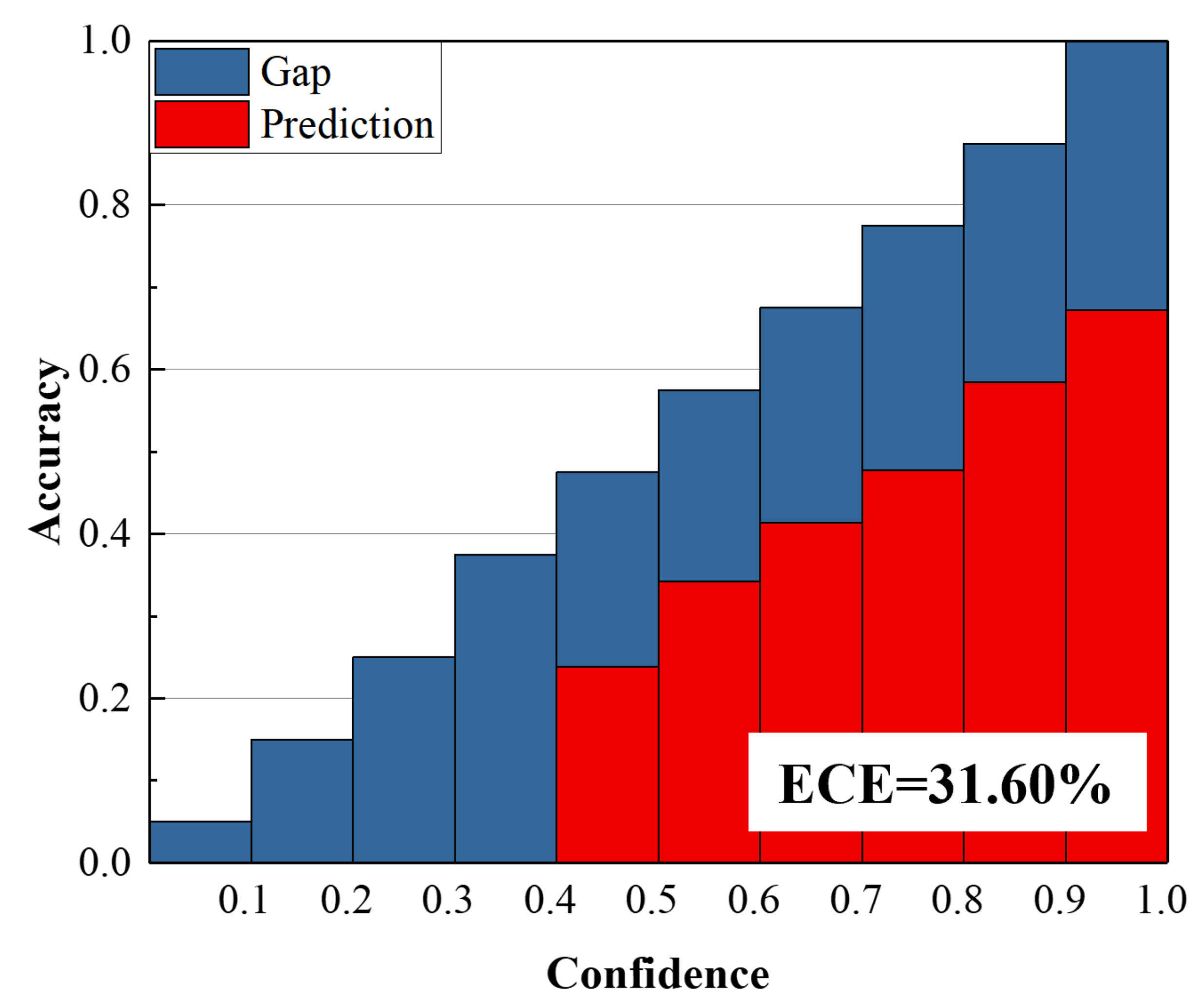}
			\end{minipage}
		}
		\subfloat[\label{tab:ece_fcn16}]{
			\begin{minipage}[t]{0.32\textwidth}
				\centering
				\includegraphics[width=\textwidth]{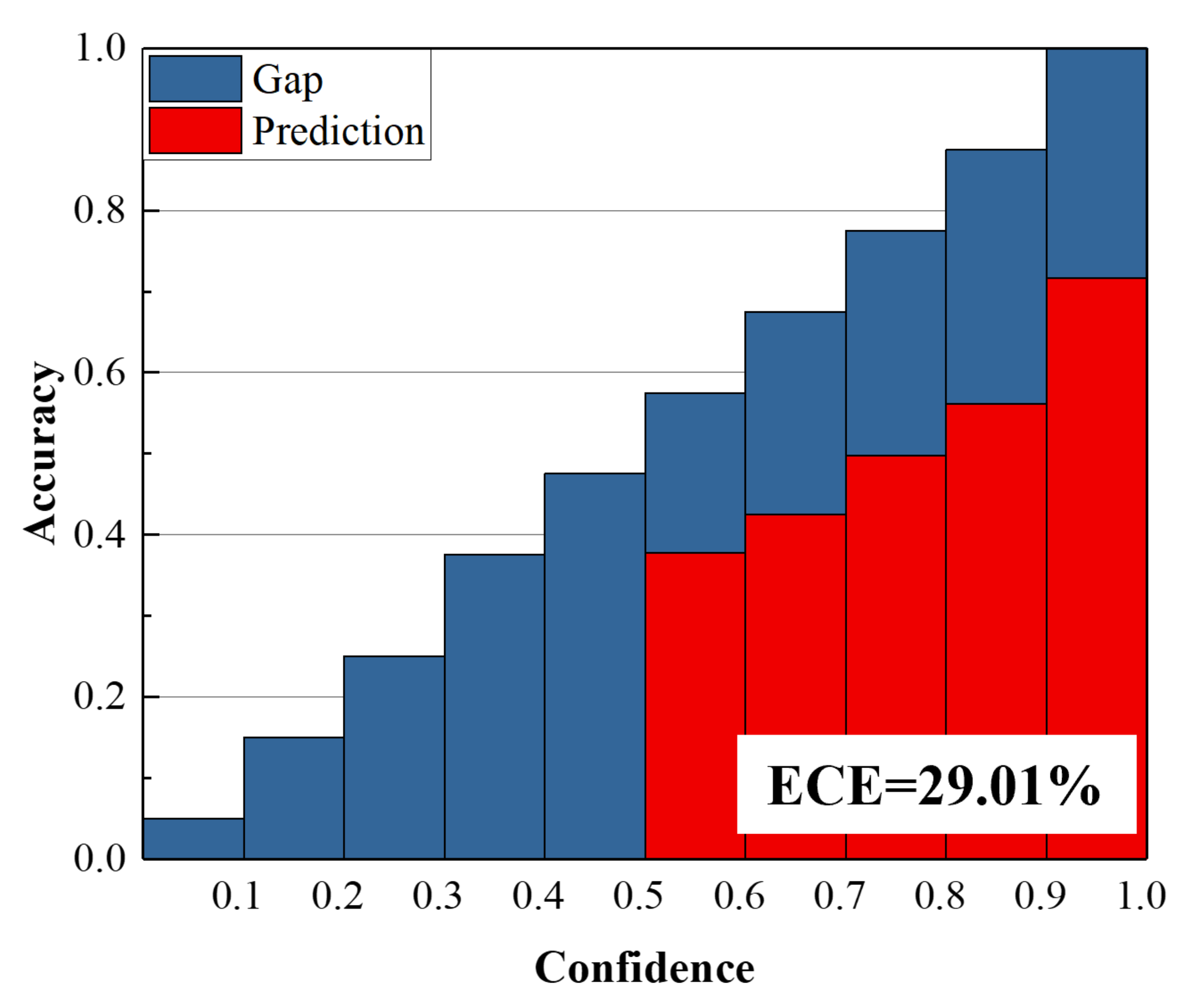}
			\end{minipage}
		}
		\subfloat[\label{tab:ece_fcn8}]{
			\begin{minipage}[t]{0.32\textwidth}
				\centering
				\includegraphics[width=\textwidth]{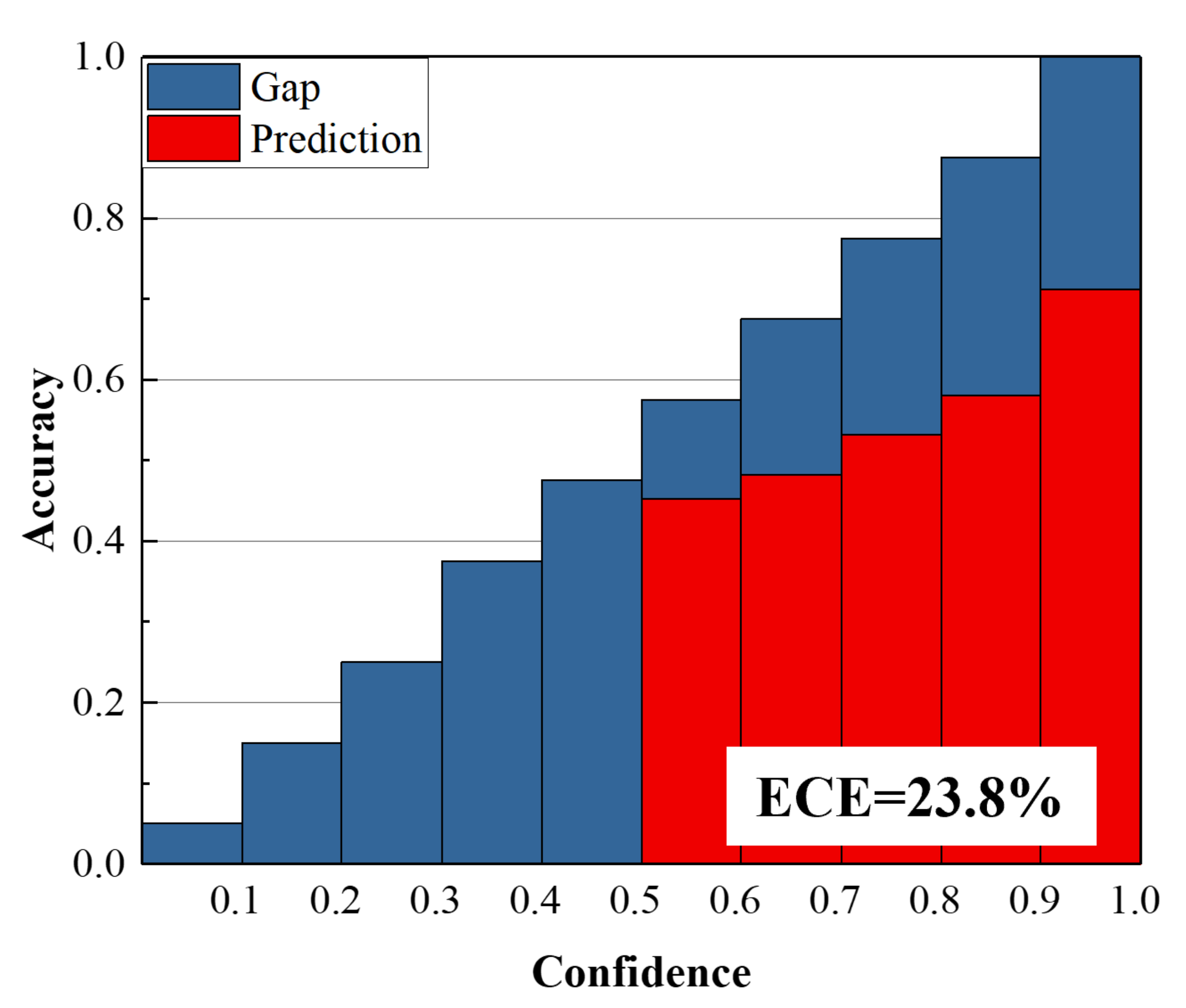}
			\end{minipage}
		}\\
		\subfloat[\label{tab:ece_unet}]{
			\begin{minipage}[t]{0.32\textwidth}
				\centering
				\includegraphics[width=\textwidth]{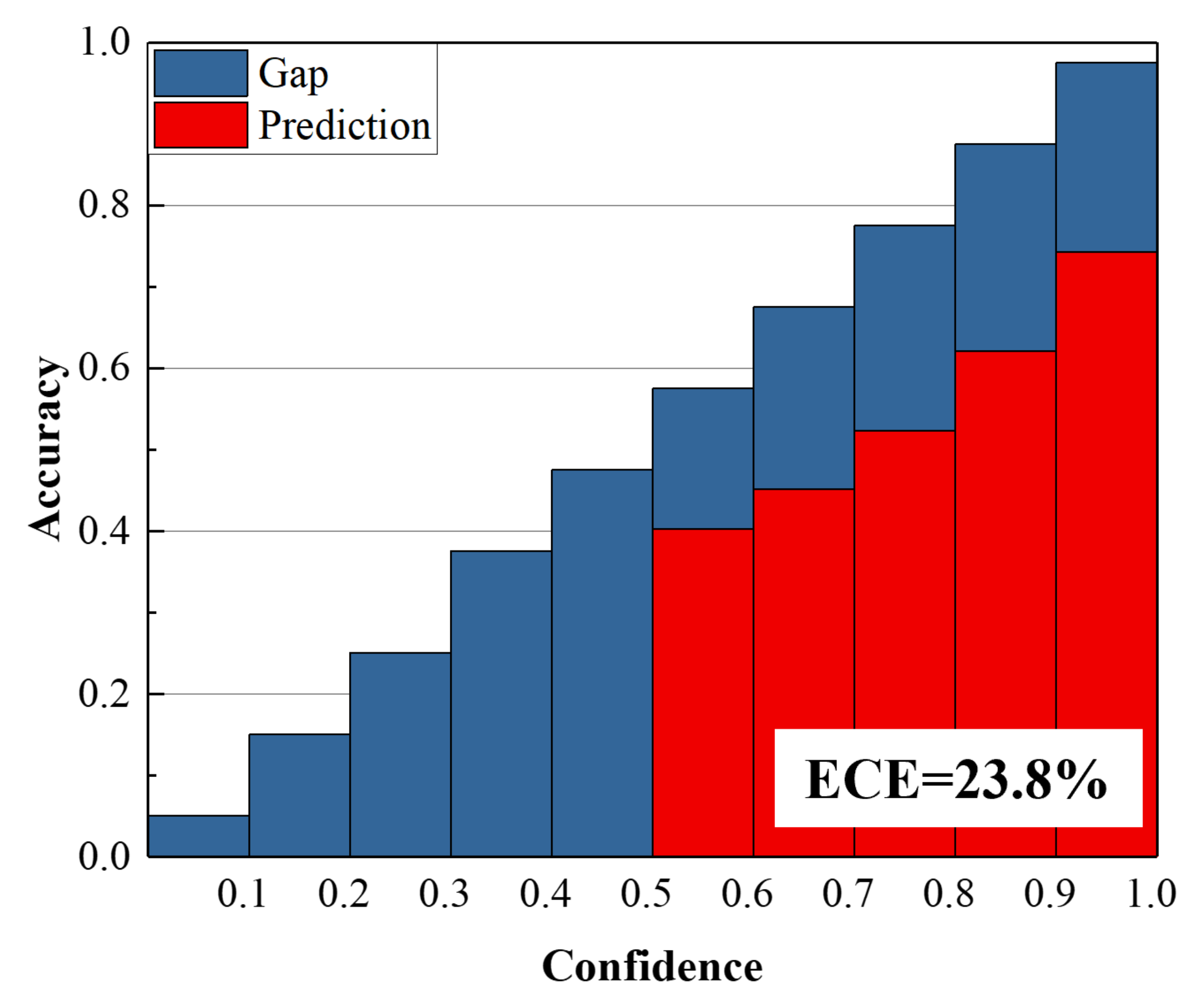}
			\end{minipage}
		}
		\subfloat[\label{tab:ece_deeplab}]{
			\begin{minipage}[t]{0.32\textwidth}
				\centering
				\includegraphics[width=\textwidth]{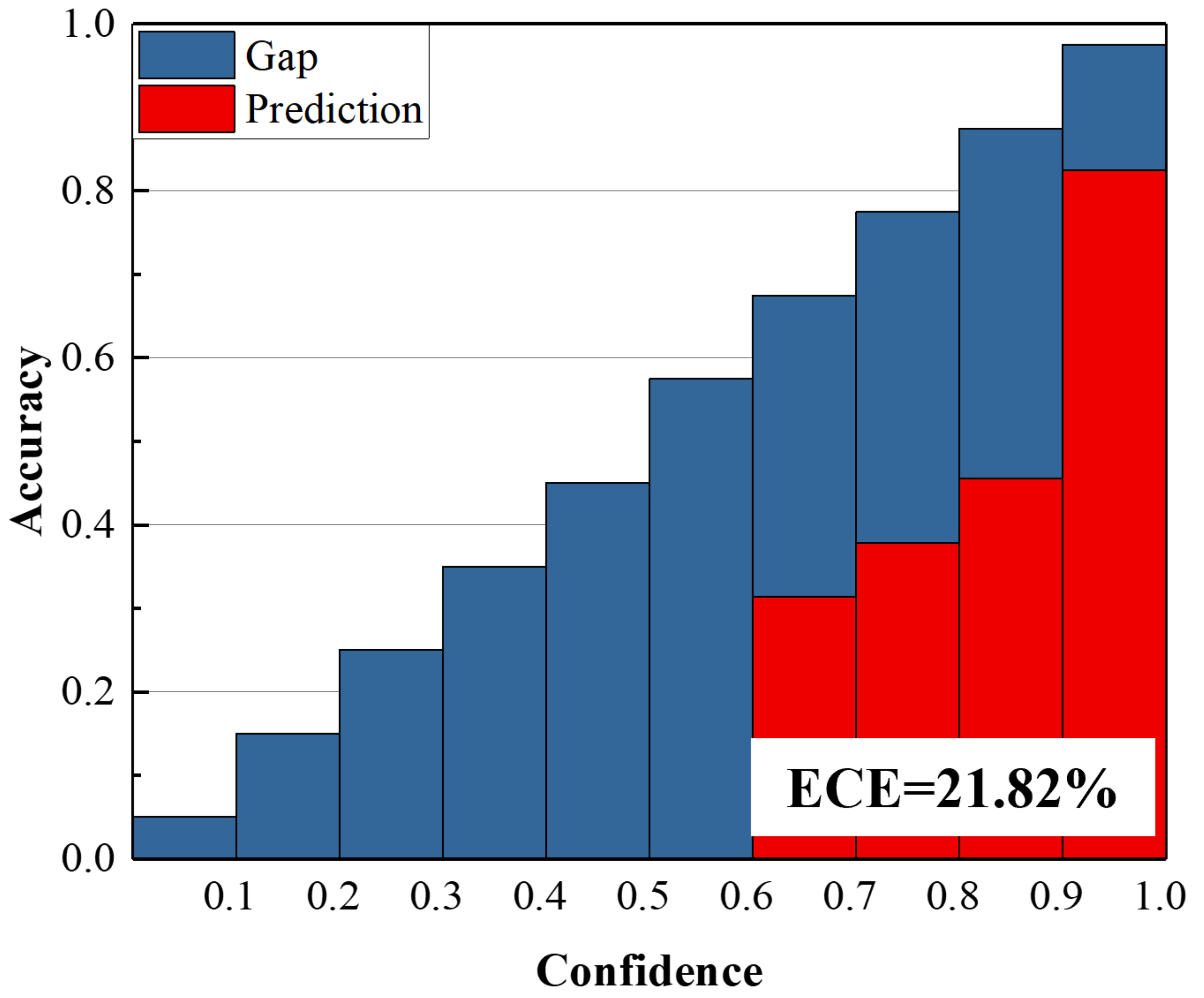}
			\end{minipage}
		}
		\subfloat[\label{tab:ece_self_attention}]{
			\begin{minipage}[t]{0.32\textwidth}
				\centering
				\includegraphics[width=\textwidth]{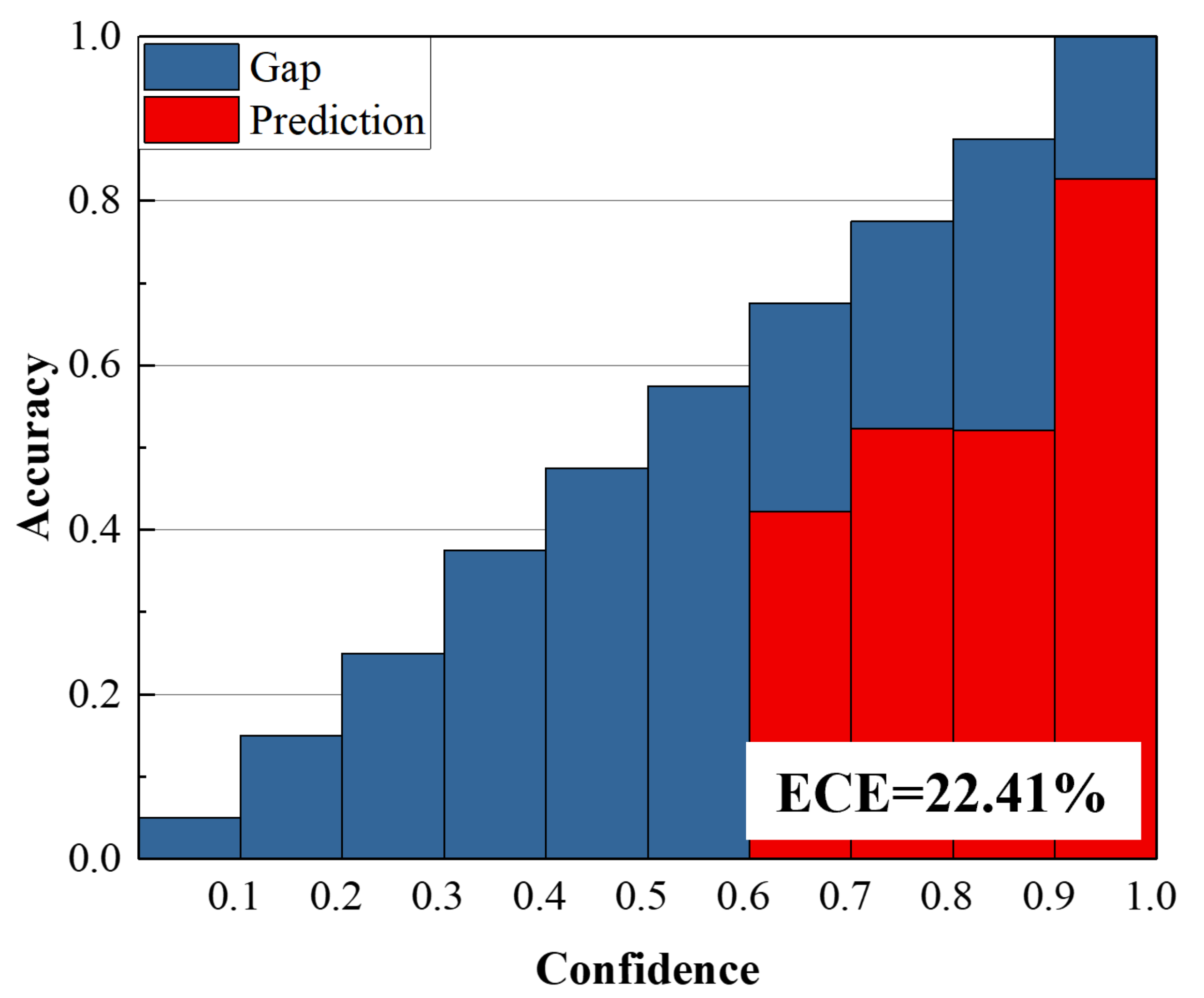}
			\end{minipage}
		}\\
		\subfloat[\label{tab:ece_cc_attention}]{
			\begin{minipage}[t]{0.32\textwidth}
				\centering
				\includegraphics[width=\textwidth]{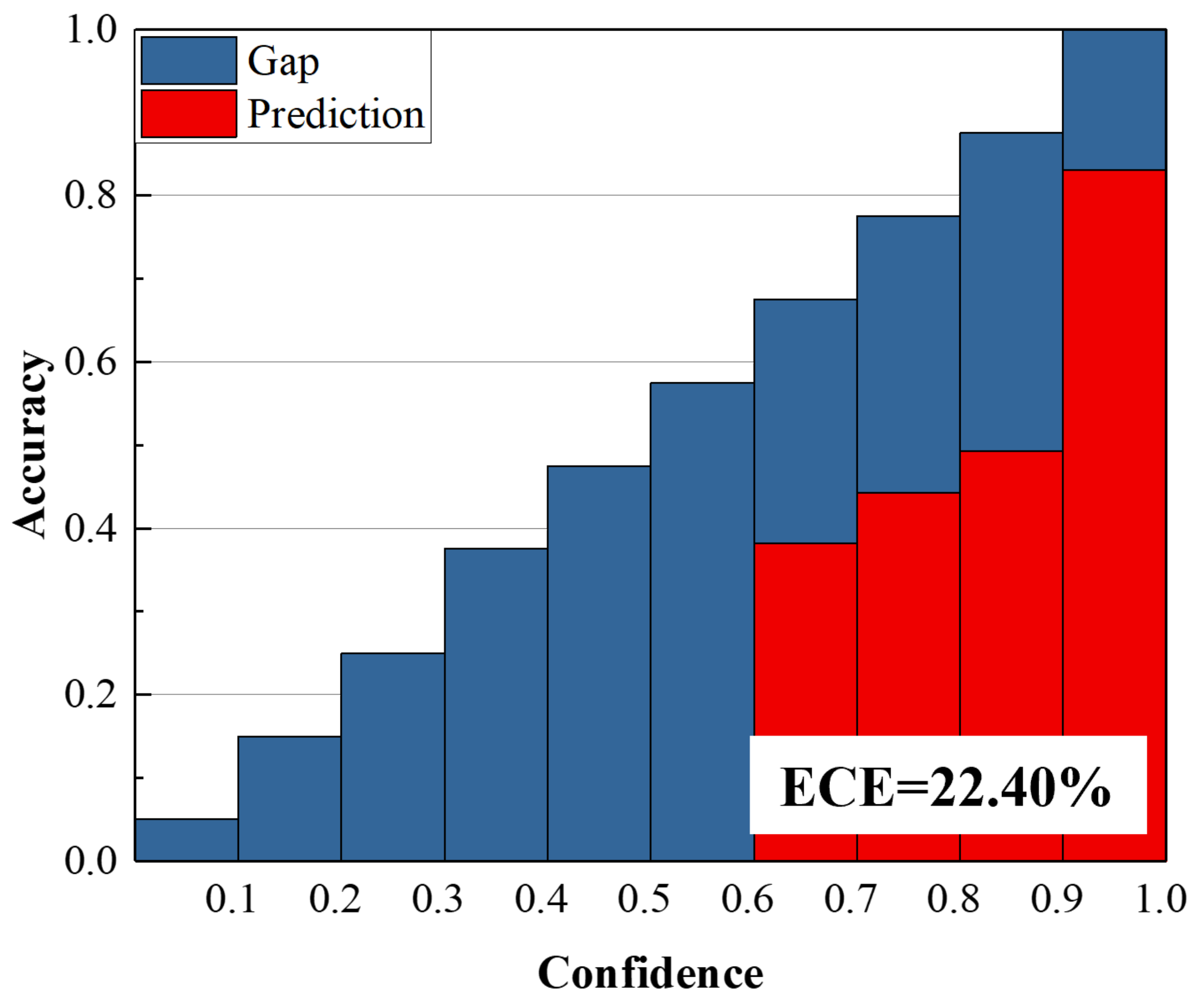}
			\end{minipage}
		}
		\subfloat[\label{tab:ece_double_attention}]{
			\begin{minipage}[t]{0.32\textwidth}
				\centering
				\includegraphics[width=\textwidth]{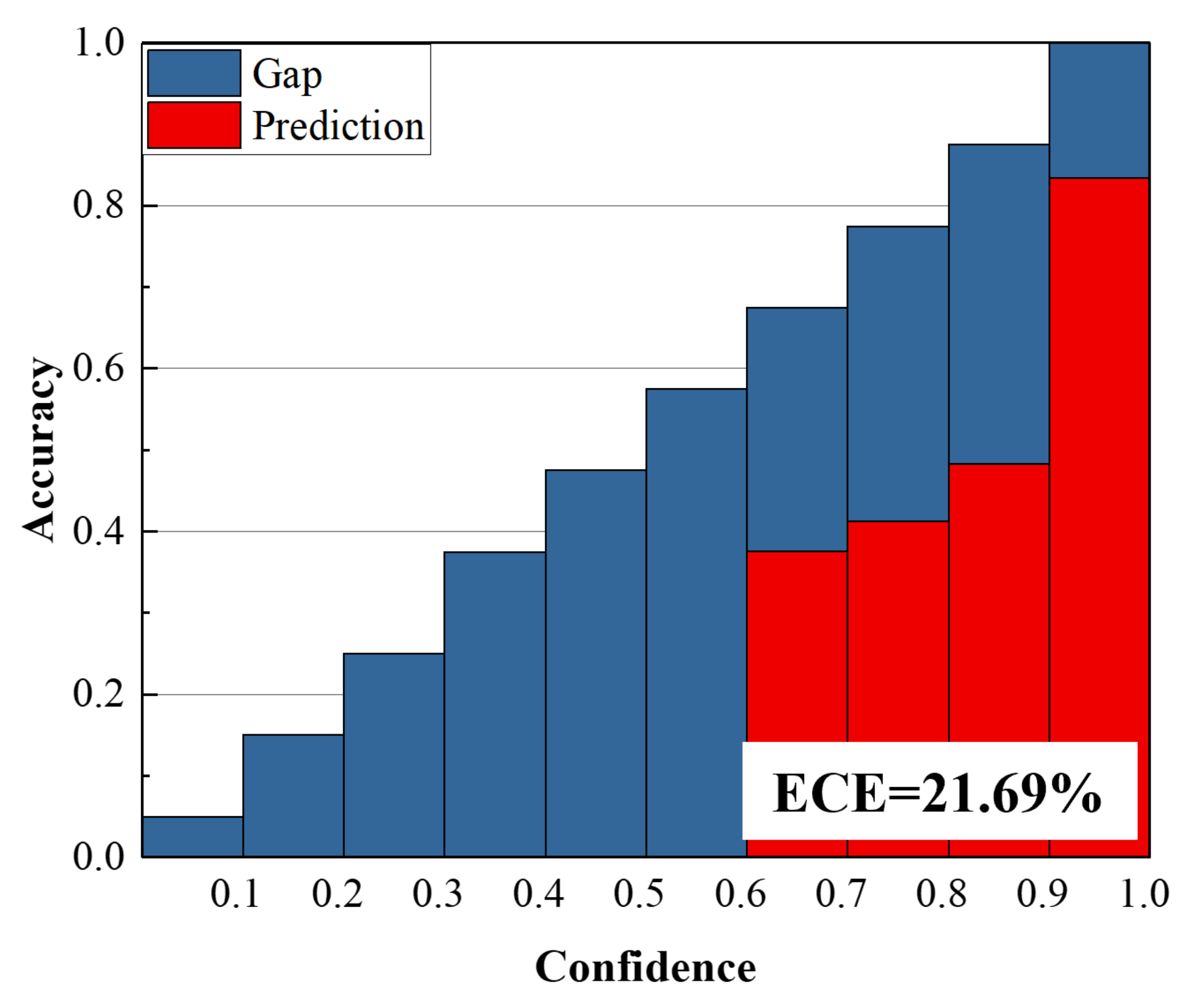}
			\end{minipage}
		}
		\subfloat[\label{tab:ece_transformer}]{
			\begin{minipage}[t]{0.32\textwidth}
				\centering
				\includegraphics[width=\textwidth]{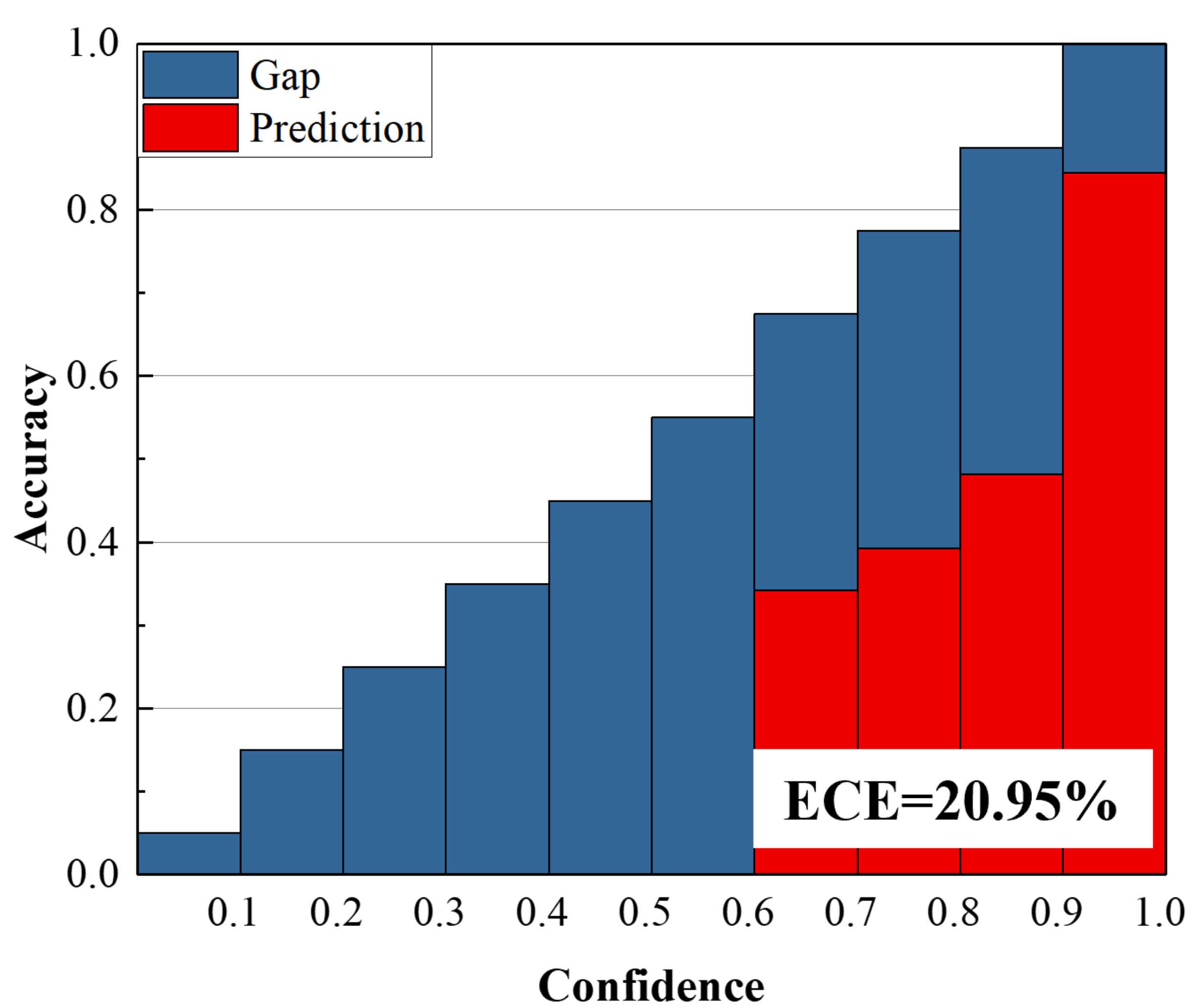}
			\end{minipage}
		}\\
		\caption{Testing pixel accuracy histograms on the Pavementscapes dataset: (a) FCN-32s, (b) FCN-16s, (c) FCN-8s, (d) U-net, (e) DeepLabv3+, (f) Self-attention net, (g) CC-attention net, (h) Double-attention net, (i) Segmentation transformer.}\label{fig:ece}
	\end{figure}

    Table \ref{tab:overall_testing} also shows that the two types of deep neural networks have similar ECEs, demonstrating the two types of deep neural networks are over-confident because their accuracies do not match their confidences. Figures \ref{fig:pixel_distribution} and \ref{fig:ece}, respectively, shows the pixel distribution and pixel accuracy histograms of the deep neural networks in Table \ref{tab:overall_testing}. Note that the background pixels are not considered in the two figures. The average confidence of each network is substantially higher than its average pixel accuracy, indicating that the network is not calibrated. This problem is mainly because the deep neural networks work within the probabilistic framework, in which the features from the backbone are imported into a softmax layer to generate probabilities of the classes for decision-making. Probability theory only captures the randomness aspect of the features but neither ambiguity nor incompleteness \cite{jiao2015hybrid,jiao2015belief}, which are inherent in damage features. For example, a deep neural network may extract incomplete damage features from an image with a non-iconic view. Besides, a network sometimes extracts ambiguous features from some small damages. Such uncertain features lead that multiple classes having similar probabilities. In such a case, deep neural networks in the probabilistic framework often arbitrarily assign the pixel to one and only one of the possible classes, which may result in misclassification and finally leads to over-confidence. The problem of over-confidence is common in the deep neural networks work within the probabilistic framework \cite{guo2017calibration}. Section \ref{sec:recommendationss} will provide a potential way to solve the problem.
    
    Table \ref{tab:overall_testing} indicates the conflict between performance and computation cost. Compared with the convolution-based models except for DeepLabv3+, the attention-based models cost more FLOPs but have larger PAs and mIoUs. Moreover, the training time of an attention-based model is twice that of a convolution-based model, demonstrating a small improvement in terms of PAs and mIoUs always requires large increases in computation costs. The costs of the attention-based networks are unbearable, even though the GPU memory and computation force has a significant increase in recent years. Therefore, attention-based deep neural networks with light weights are required for the task of pavement damage segmentation.
    
\begin{figure}[htbp]
    	\centering
    	\subfloat[\label{tab:stability_service_year}]{
    		\begin{minipage}[t]{\textwidth}
    			\centering
    			\includegraphics[width=\textwidth]{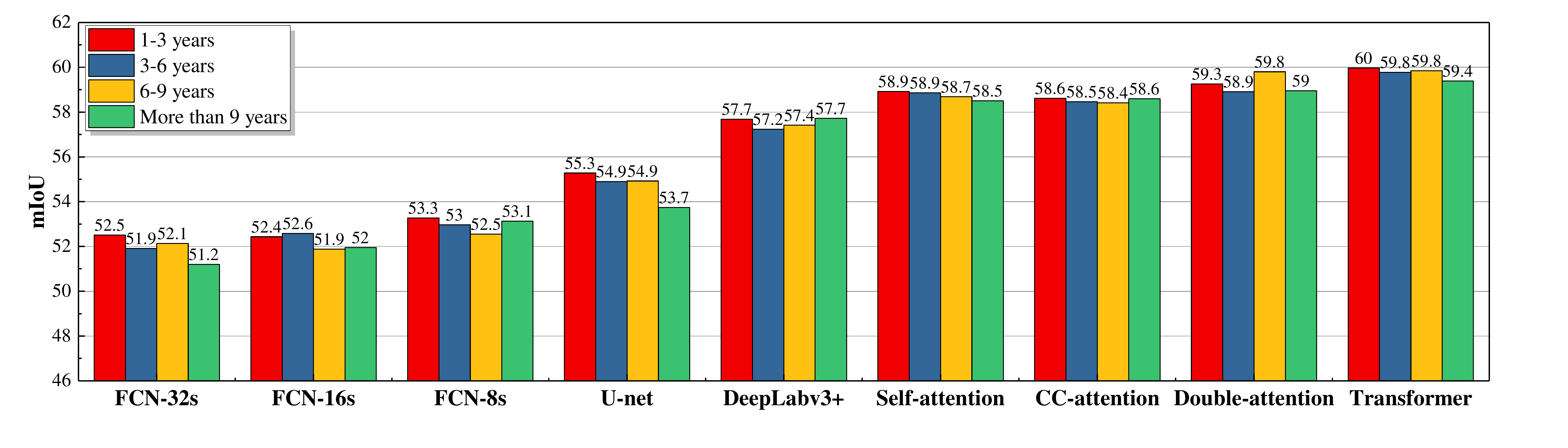}
    		\end{minipage}
    	}\\
    	\subfloat[\label{tab:stability_weather}]{
    		\begin{minipage}[t]{\textwidth}
    			\centering
    			\includegraphics[width=\textwidth]{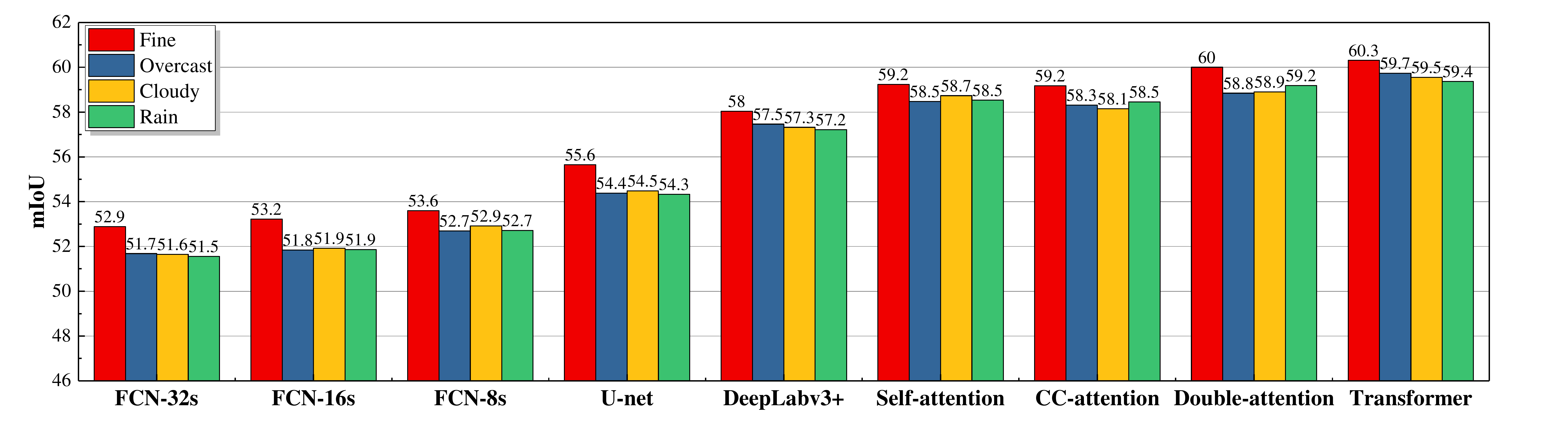}
    		\end{minipage}
    	}\\
    	\subfloat[\label{tab:stability_materials}]{
    		\begin{minipage}[t]{\textwidth}
    			\centering
    			\includegraphics[width=\textwidth]{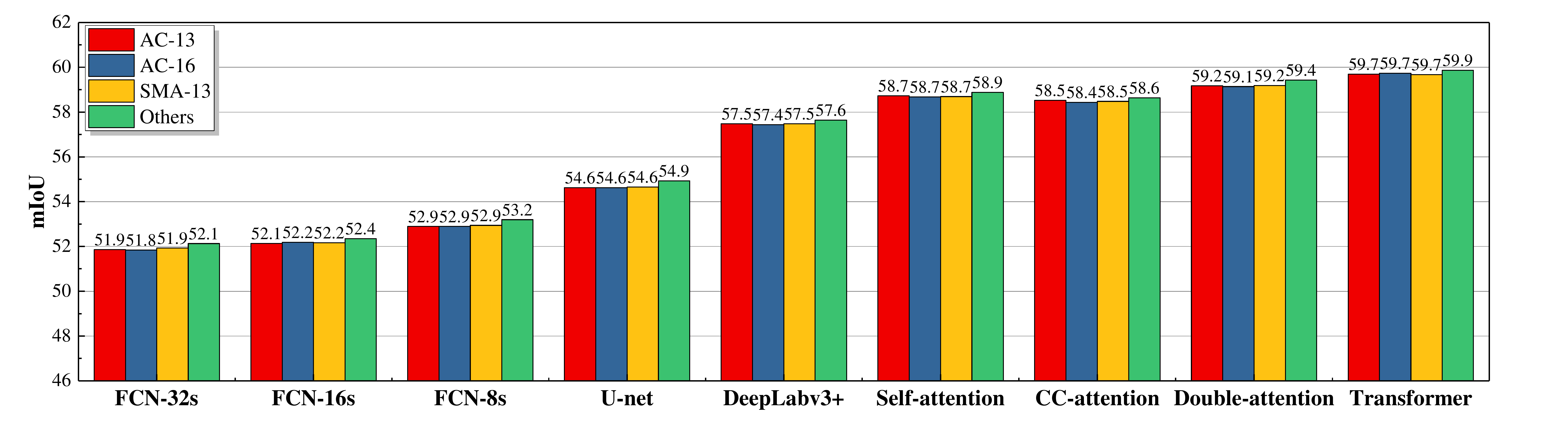}
    		\end{minipage}
    	}\\
    	\caption{Stability anaylsis using the Pavementscapes dataset under different (a) service years, (b) weathers, and (c) pavement materials.}\label{fig:stability}
    \end{figure}

    Figure \ref{fig:stability} presents the mIoUs of the two types of deep neural networks under various real-world conditions. These networks have stable performances on different service years, weathers, and pavement materials since their IoUs and PUs do not significantly change under different conditions. This demonstrates that the two types of deep neural networks can perform damage segmentation well in the real world.
    
	\subsection{Recommendations and future scopes}
	\label{sec:recommendationss}
	
	Deep neural networks achieve some good results on the Pavementscapes dataset, which can be considered as the baseline of the PavmentScapes challenges. However, they still face several problems that are not easy to be solved by using traditional deep neural networks. This study provides the recommendations for these problems as follows.
	\begin{enumerate}
		\item \emph{Small damage instances.} Current deep neural networks sometimes ignore cracks with thin widths, especially the convolution-based networks, though the features of these cracks are important for decision-making. This behavior makes these networks have low performances on crack segmentation. Attention-based deep neural networks show the potential capacity to solve this problem. In the future, more attention-based networks should be trained and tested by the Pavementscapes dataset, which may replace the convolution-based networks for pavement damage segmentation.
		
		\item \emph{Unbalanced training set.} Even though the Pavementscapes dataset has the real-world distribution of different damage instances, it still has the unbalanced problems in the numbers of different classes, such that the background pixels is much more than the sum of the damage pixels and the crack pixels are less than the pixels of other damages. This fact introduces a negative effect on learning systems that tend to assign a pixel to the background class or the damage classes with a large number of pixels. Similar behaviors are common in the segmentation task of medical and cell images \cite{craddock2013neuro,levy2016breast}. In the medical image segmentation, many morphology-based loss functions are used to solve the unbalanced problem, such as focal loss \cite{lin2017focal}, dice loss \cite{sudre2017generalised} used in the experiment, and IoU Loss \cite{zheng2020distance}. Therefore, these loss functions should be introduced into the deep neural networks to improve the performance of crack segmentation.
		
		\item \emph{Over-confidence.} Two types of deep neural networks are not calibrated well in the damage segmentation task. This problem derives from the use of the probability framework.  During the last decade, many theories have been combined with deep neural networks to solve the problem, and one of the successful cases is the evidential deep neural network \cite{tong2019convnet}, which converts the features from the backbone of a deep neural network into Dempster-Shafer belief functions, rather than the probabilities using a softmax layer. This architecture allows the network to represent the feature uncertainty \cite{tong2021evidentialcnn} and reduce the confidence of the network \cite{tong2021evidentialfcn}. Such architecture should be considered to be combined with the attention-based models to make the deep neural networks well-calibrated.
		
	\end{enumerate}
	
\section{Conclusions}
\label{sec:conclusions}

This study has proposed a large-scale hierarchical image dataset for asphalt pavement damage segmentation, called Pavementscapes. The statistical study and the deep learning experiment provide an in-depth analysis of the dataset. The following conclusions are can be drawn.

\begin{enumerate}
	\item The Pavementscapes dataset consists of 4,000 pavement images with a resolution of $1024\times2048$ and 8,680 damage instances, which were recorded from several real-world projects of pavement inspection in China. Six damage classes are included in the dataset. The proposed dataset exceeds the other public pavement dataset in the number of pavement images, damage classes, annotation levels, and shooting views.
	\item The statistical analysis demonstrates that the Pavementscapes dataset has a reasonable damage distribution, complex pavement scenes, and high annotation accuracy, which ensure the completeness and comprehensiveness of the dataset. In addition, the images with different non-iconic views improve the complexity and truth of the dataset. In summary, the dataset represents the real-world pavement damages well.
	\item The numerical experiment uses the Pavementscapes dataset to train and test the top-performance deep neural networks. The results demonstrate that the deep neural networks have the powerful potential to segment pavement damages once given enough good training samples. The attention-based models outperform the convolution-based ones on the segmentation task, which may be the new direction for visual damage segmentation. The experiment results can be considered as the baseline for the public damage segmentation challenge.
	\item The results of the numerical experiment indicate three problems with the use of deep neural networks in pavement damage segmentation: the segmentation of small damage instances, the unbalanced training set, and the over-confidence of deep neural networks. The three problems are not easy to solve using the current state-of-the-art deep networks.
	\item Future work will focus on three main aspects, corresponding to the above three problems. First, the attention mechanism should be further applied in the segmentation of small damage instances, which has the potential to improve the performance of crack segmentation. Other advanced morphology-based loss functions should be introduced into deep neural networks to solve the problem of unbalanced learning set. Finally, some uncertainty frameworks, such as Dempster-Shafer theory, should be used to overcome the over-confidence problem. 
\end{enumerate}



\section*{Author Contributions}

Zheng Tong: conception and study design, data collection and analysis and interpretation, drafting the article; Tao Ma: conception and study design, reviewing the article; Ju Huyan: data collection and reviewing the article.

\section*{Acknowledgment}

This research was supported by National Key Research and Development Project (grant number 2020YFA0714302) and National Key Research and Development Project (grant number 2020YFB1600102).

\bibliography{mybibfile}

\appendix
\section{Examples of segmentation results}
\label{sec:appendix_segmentor}
There are eight examples in the appendix. For each example, Figures (a), (b), (c), and (d) are the original image, ground trurh, segmentation results from the DeepLabv3+, and segmentation results from the segmentation transformer, respectively. The masks with gray-scale values of 0 are the ``background'' pixels; other masks with different gray-scale values are the pixels belonging to different classes, such that 30, 60, 90, 120, 150, and 180 gray-scale values stand for the pixels of ``longitudinal crack'', ``lateral crack'', ``alligator crack'', ``pothole'', ``rut'', and ``repair area'', respectively.

	\begin{figure}
		\centering
		\includegraphics[width=\textwidth]{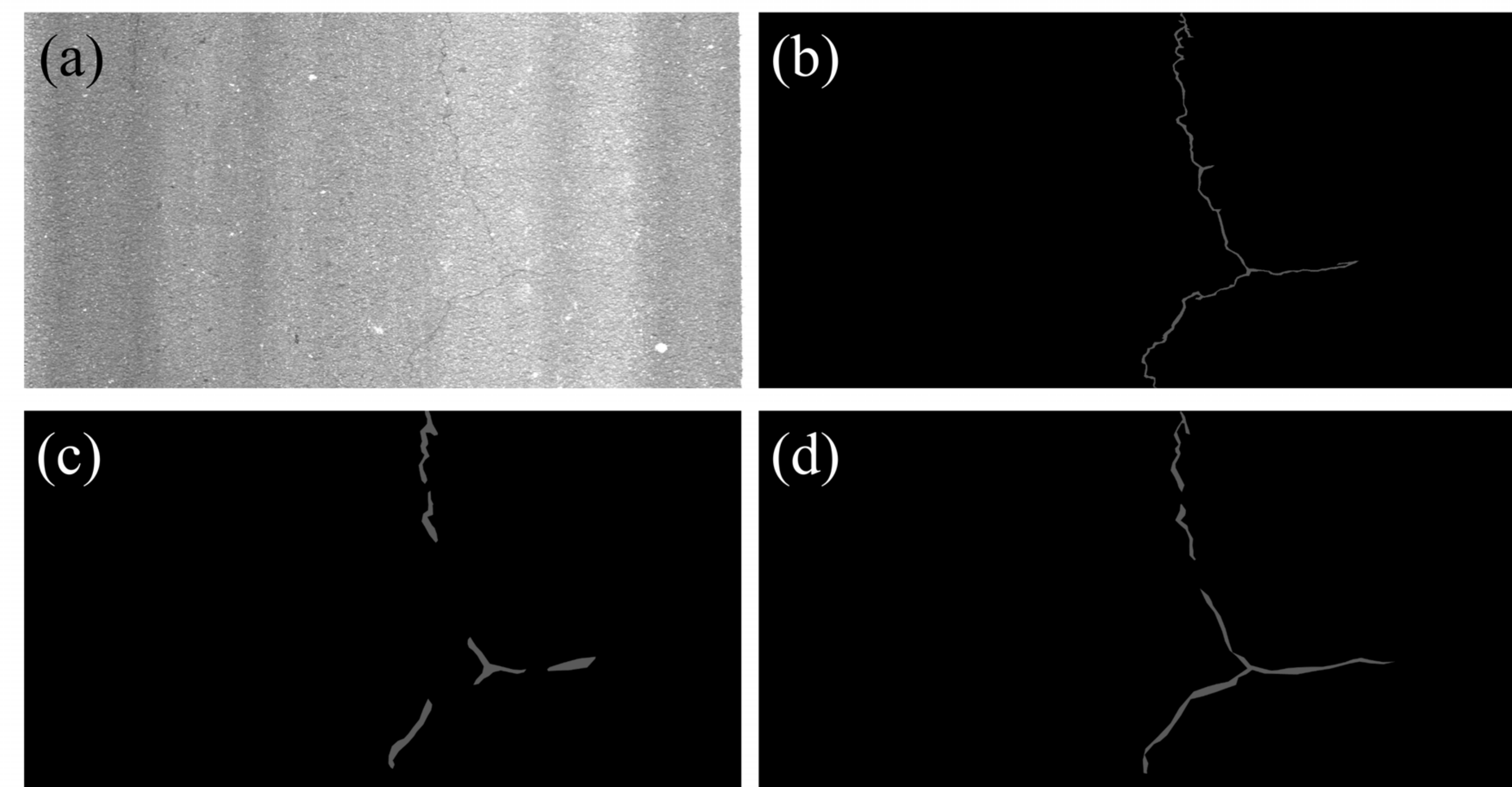}
		\caption{Example 1.}
	\end{figure}

	\begin{figure}
		\centering
		\includegraphics[width=\textwidth]{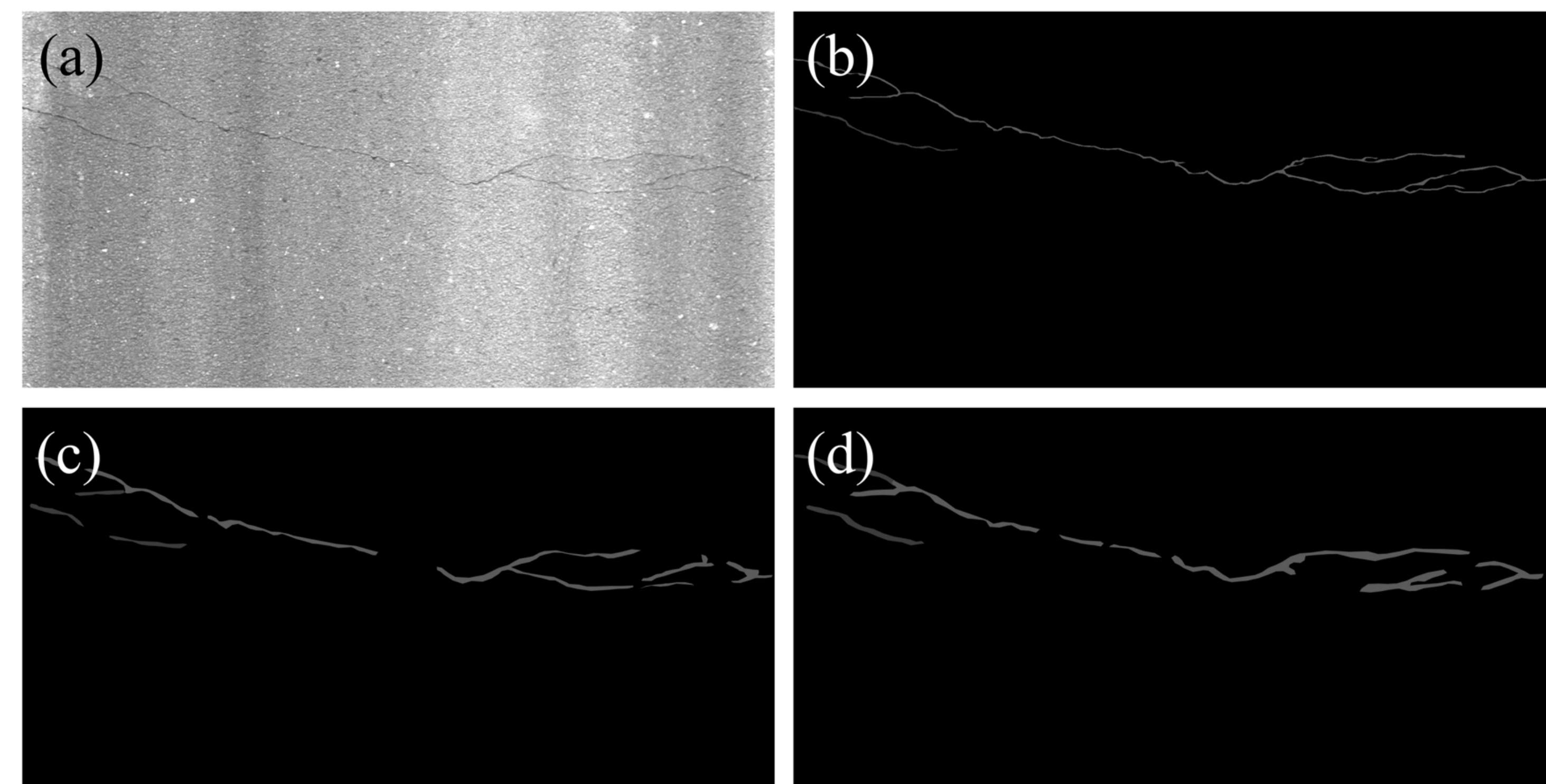}
		\caption{Example 2.}
	\end{figure}

	\begin{figure}
		\centering
		\includegraphics[width=\textwidth]{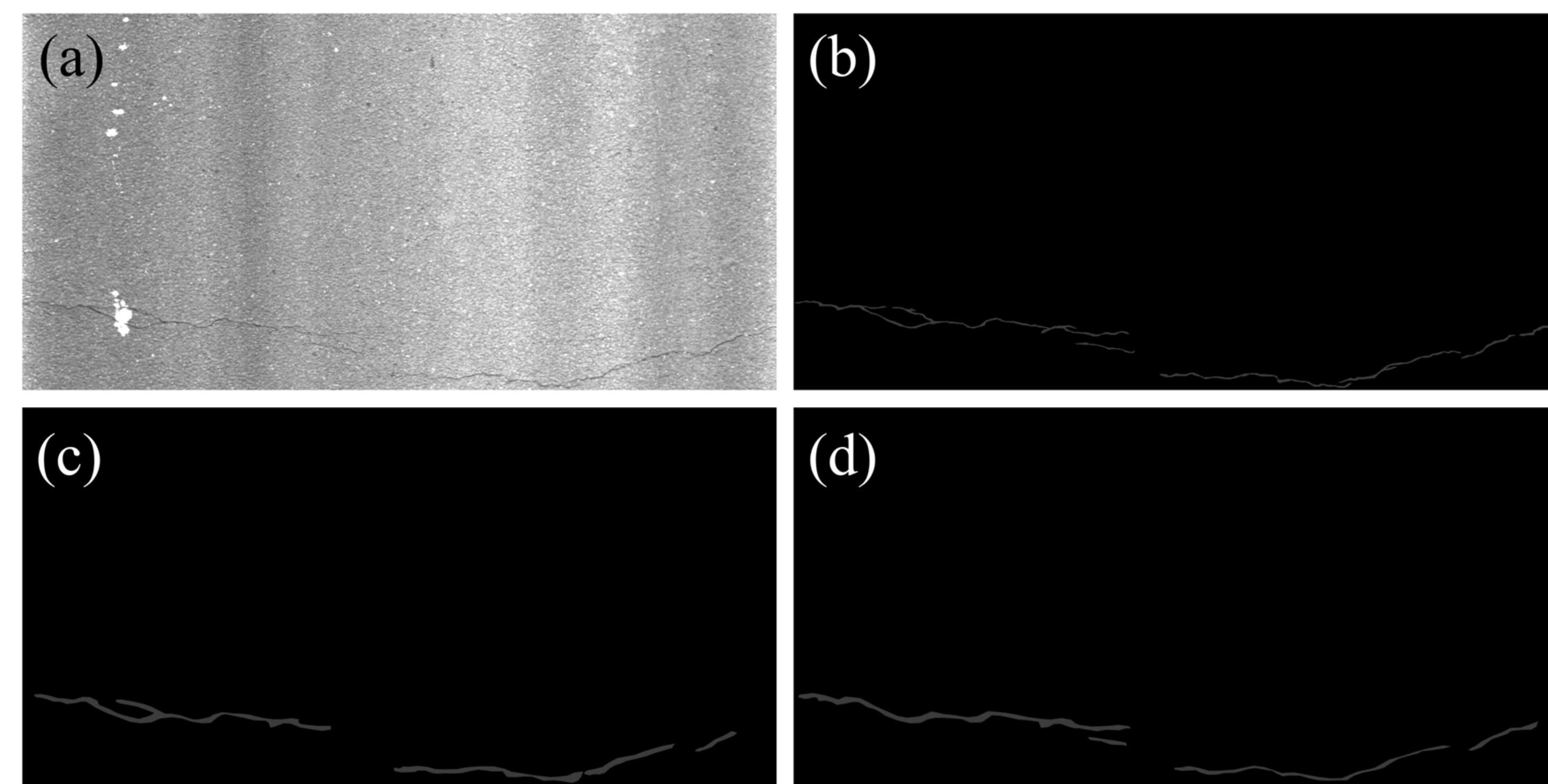}
		\caption{Example 3.}
	\end{figure}

	\begin{figure}
		\centering
		\includegraphics[width=\textwidth]{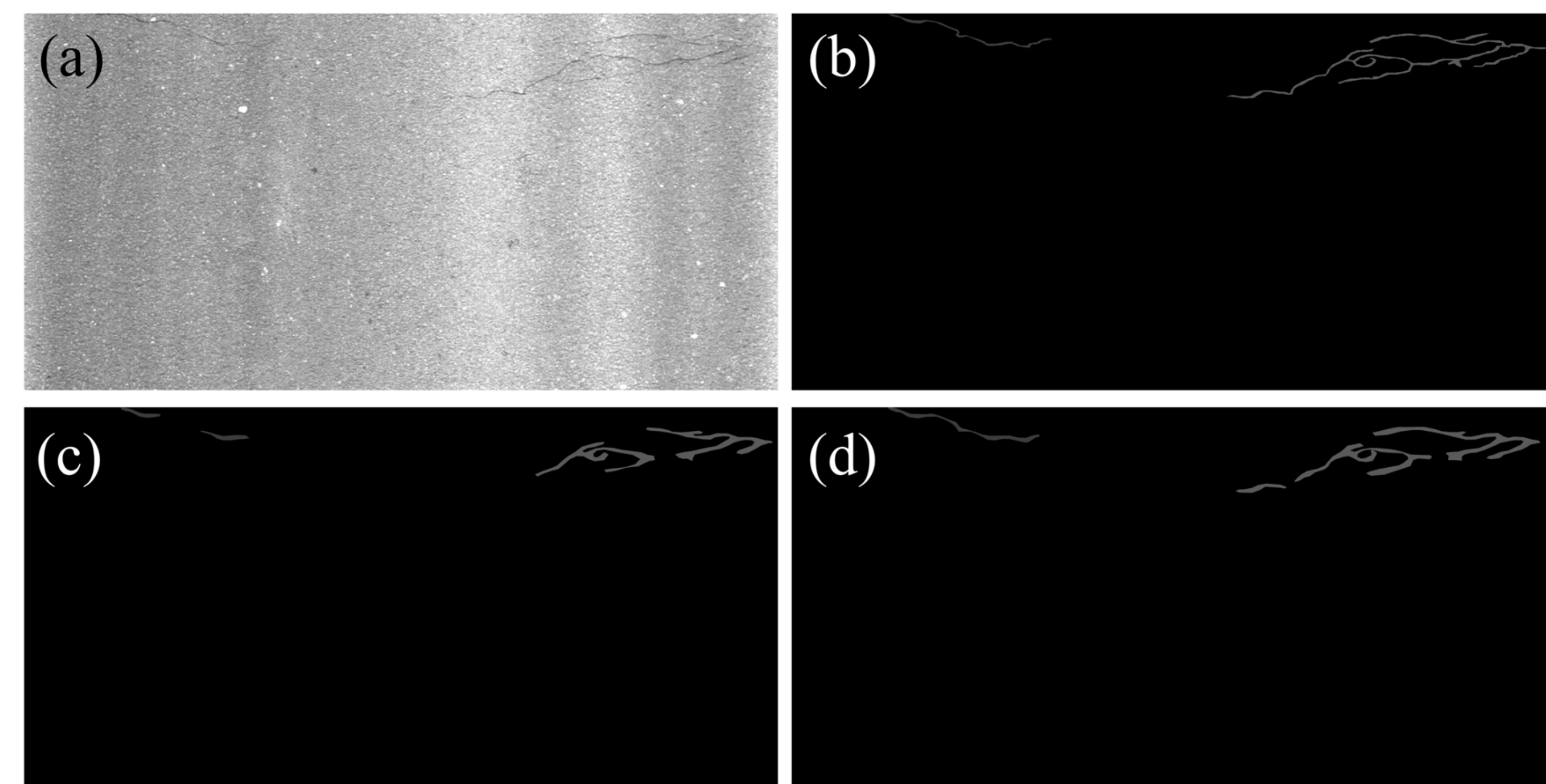}
		\caption{Example 4.}
	\end{figure}

	\begin{figure}
		\centering
		\includegraphics[width=\textwidth]{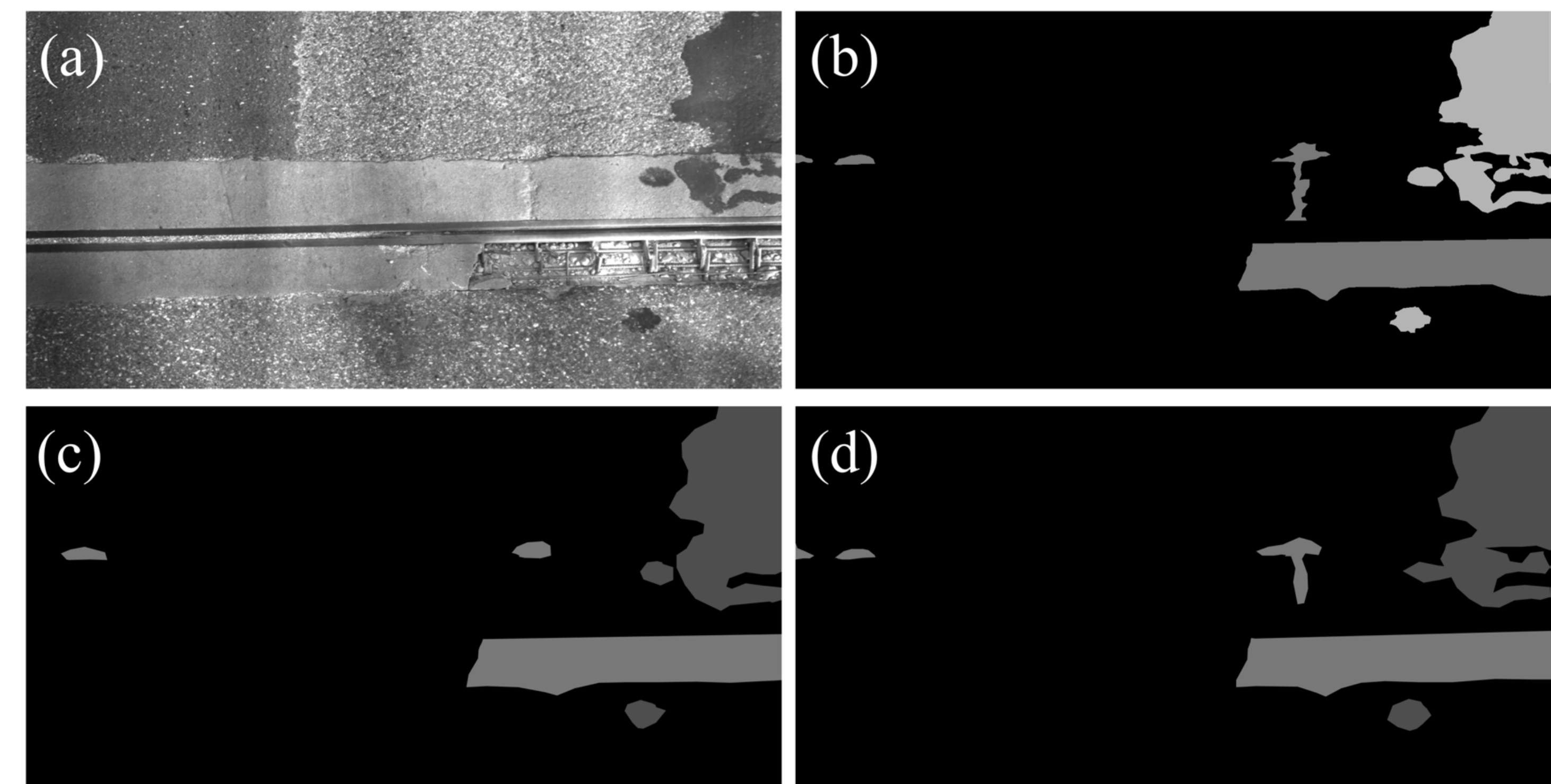}
		\caption{Example 5.}
	\end{figure}

	\begin{figure}
		\centering
		\includegraphics[width=\textwidth]{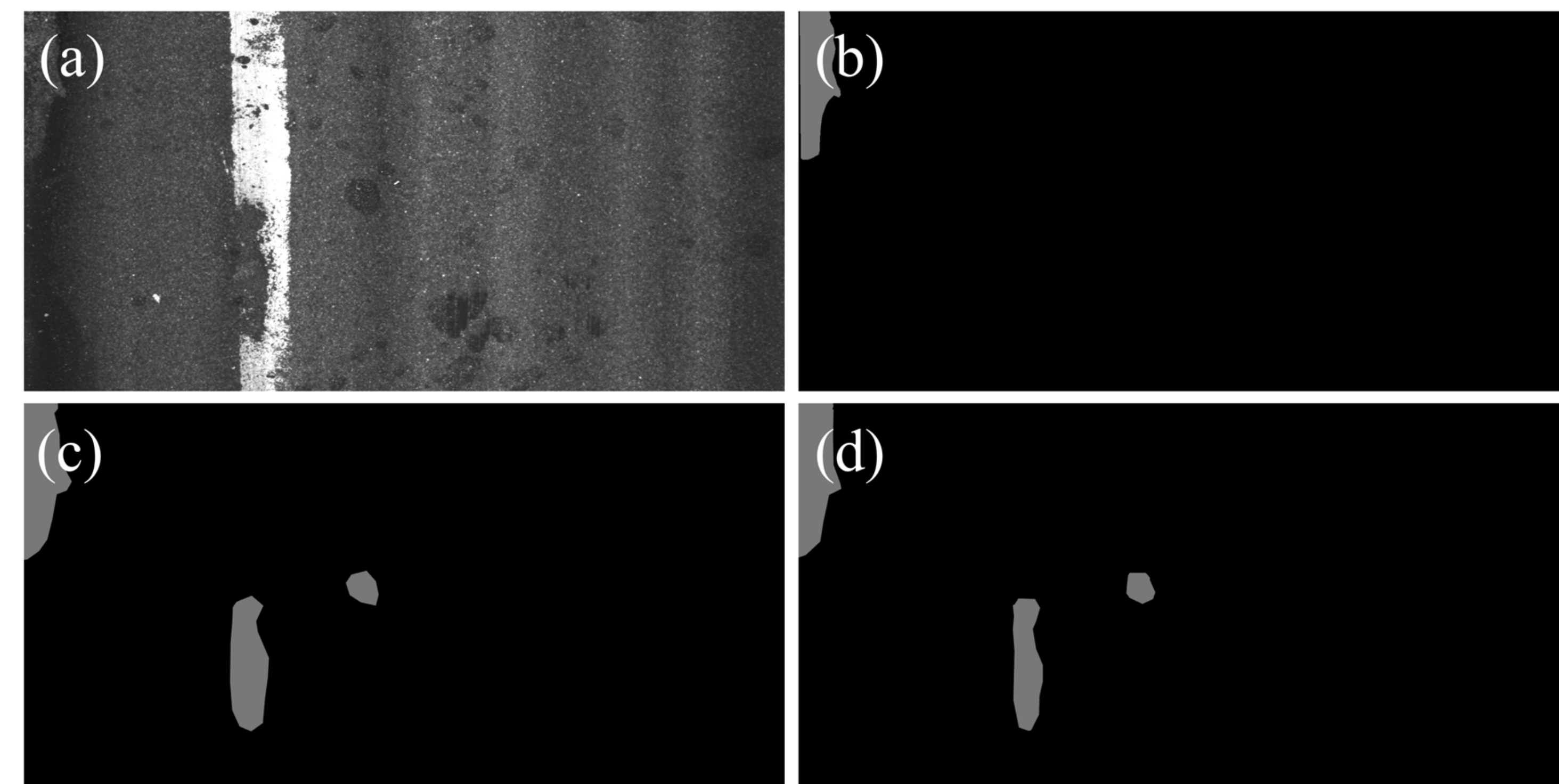}
		\caption{Example 6.}
	\end{figure}

	\begin{figure}
		\centering
		\includegraphics[width=\textwidth]{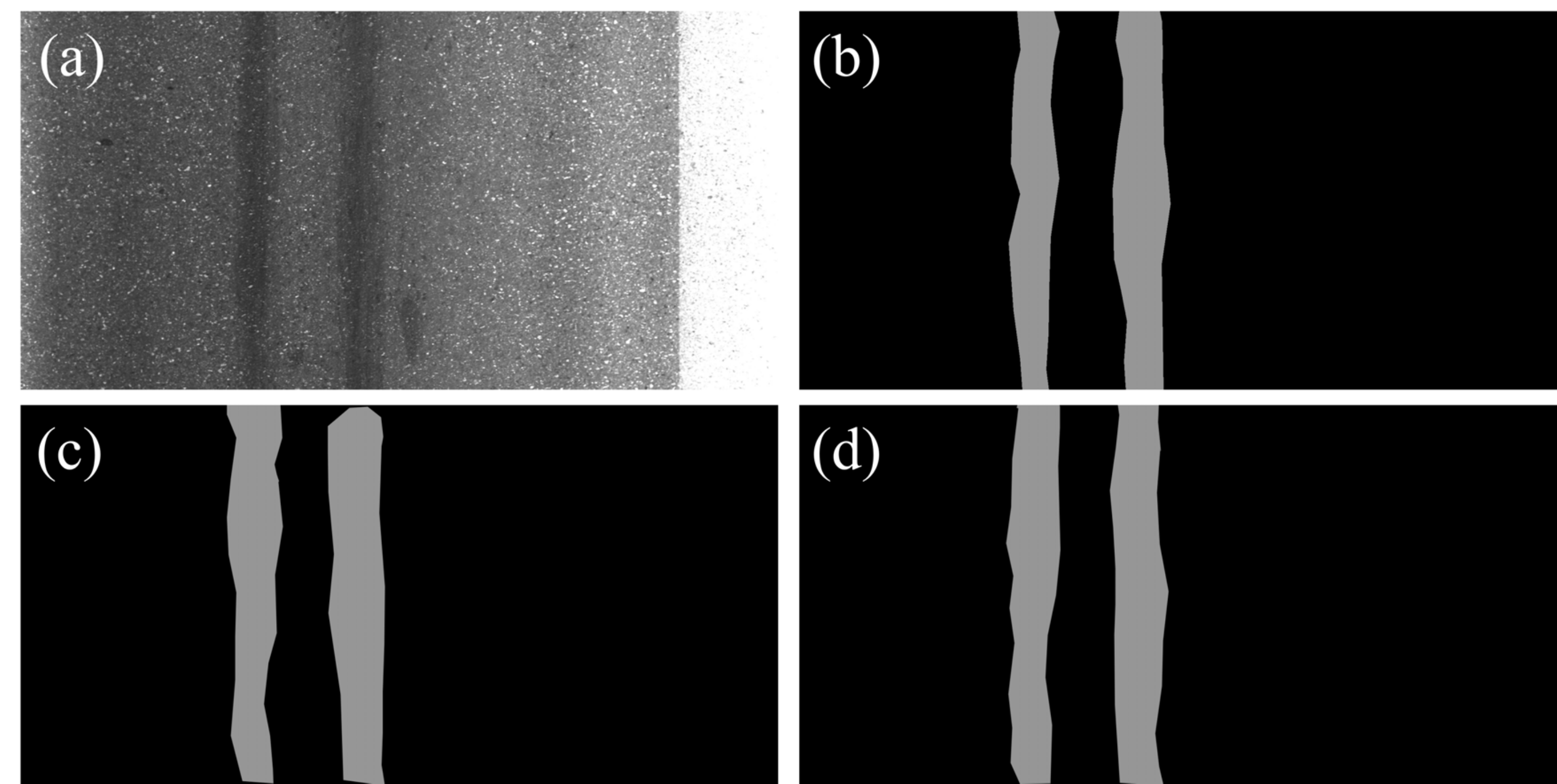}
		\caption{Example 7.}
	\end{figure}

	\begin{figure}
		\centering
		\includegraphics[width=\textwidth]{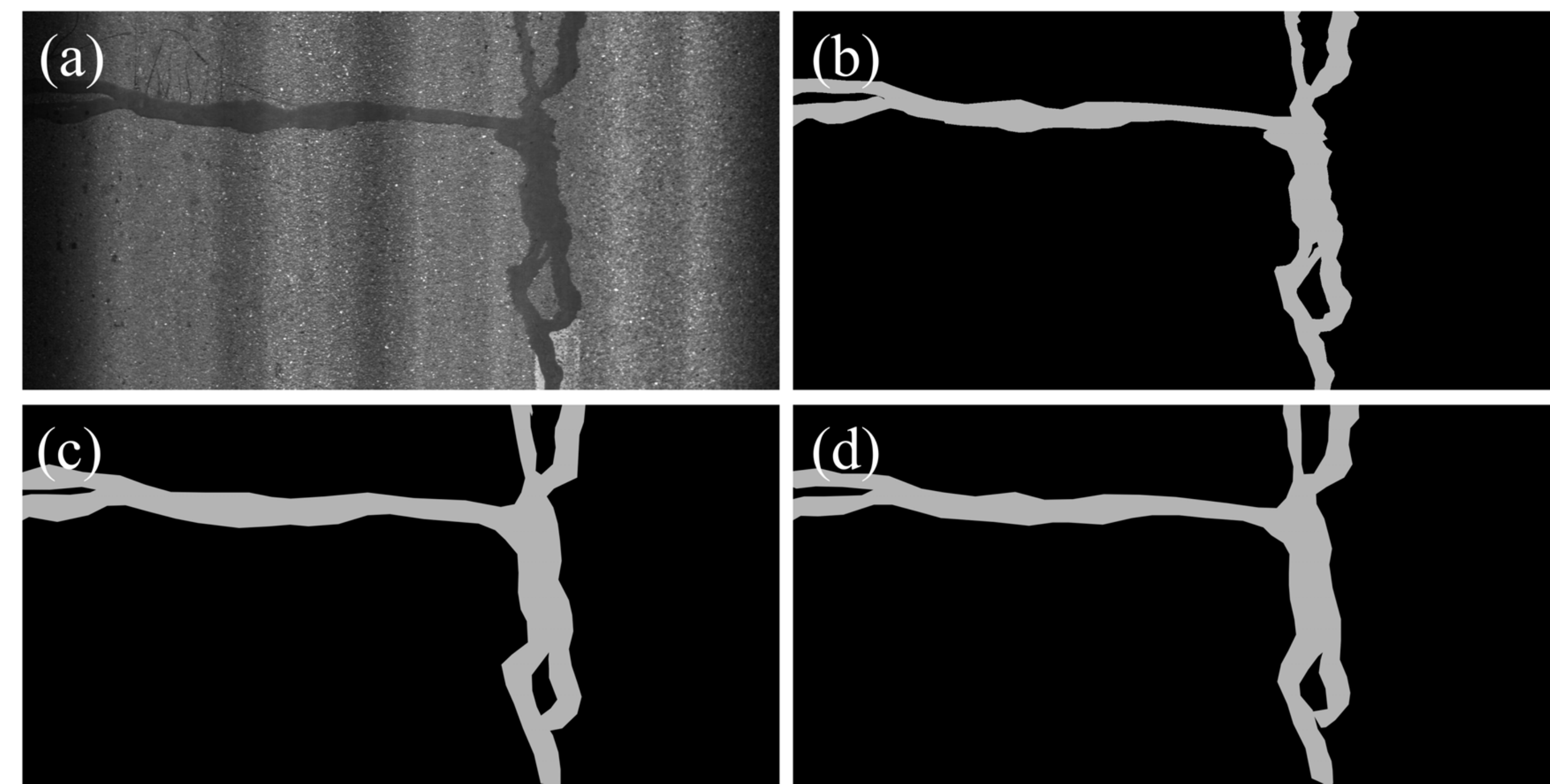}
		\caption{Example 8.}
	\end{figure}


\end{document}